\documentclass[11pt]{article}

\usepackage{acl}

\usepackage{times}
\usepackage{latexsym}
\usepackage[T1]{fontenc}
\usepackage[utf8]{inputenc}
\usepackage{microtype}
\usepackage{inconsolata}
\usepackage{fvextra}
\usepackage{graphicx}
\usepackage{booktabs}
\usepackage{array}
\usepackage{multirow}
\usepackage{arydshln}
\usepackage{amssymb}
\usepackage{amsmath}
\usepackage{xspace}
\usepackage{enumitem}
\usepackage{cleveref}

\usepackage[suppress]{color-edits}
\addauthor{vc}{red}
\addauthor{anna}{blue}
\addauthor{shanda}{orange}
\addauthor{ns}{red}

\usepackage{xcolor}
\usepackage[most]{tcolorbox}
\usepackage{fontawesome5}
\newcommand{\githubicon}{\textcolor{gray!85!black}{\faGithub}}
\newcommand{\roboticon}{\textcolor{blue!65!black}{\faRobot}}
\tcbset{
  reprobox/.style={
    enhanced,
    breakable,
    boxrule=0.5pt,
    arc=3pt,
    left=8pt,
    right=8pt,
    top=7pt,
    bottom=7pt,
    toptitle=5pt,
    bottomtitle=5pt,
    before skip=7pt,
    after skip=9pt,
    fonttitle=\bfseries\small,
    coltitle=black
  }
}
\newtcolorbox{humanissuebox}[1][]{
  reprobox,
  title={\githubicon\ Human GitHub issue},
  colback=gray!3,
  colframe=gray!55,
  colbacktitle=gray!12,
  #1
}
\newtcolorbox{agentfindingbox}[1][]{
  reprobox,
  title={\roboticon\ Agent output},
  colback=blue!2,
  colframe=blue!45,
  colbacktitle=blue!8,
  #1
}
\newcommand{\ghpill}[2][gray]{%
  \tcbox[
    on line,
    boxsep=0pt,
    left=3pt,
    right=3pt,
    top=1pt,
    bottom=1pt,
    boxrule=0.4pt,
    arc=6pt,
    colback=#1!8,
    colframe=#1!45
  ]{\scriptsize\bfseries\sffamily #2}%
}
\newcommand{\issuefield}[2]{%
  \par\smallskip\noindent{\scriptsize\bfseries\textcolor{gray!70!black}{#1}}\par
  \noindent #2\par
}

\newcommand{\ourmethod}{ReproRepo\xspace}
\newcommand{\codex}{Codex\xspace}
\newcommand{\cmark}{\ensuremath{\checkmark}}

\title{\ourmethod: Scaling Reproducibility Audits with GitHub Repository Issues}

\author{
Shanda Li$^{\spadesuit}$\qquad Qiuhong Anna Wei$^{\spadesuit}$\qquad Jingwu Tang$^{\spadesuit}$\qquad
Valerie Chen$^{\spadesuit}$\\
\textbf{Nihar B Shah}$^{\spadesuit}$\qquad \textbf{Tim Dettmers}$^{\spadesuit}$\qquad \textbf{Yiming Yang}$^{\spadesuit}$\qquad \textbf{Ameet Talwalkar}$^{\spadesuit\heartsuit}$\\
$^{\spadesuit}$School of Computer Science, Carnegie Mellon University\qquad 
$^{\heartsuit}$Datadog\\
\texttt{shandal@cs.cmu.edu}
}

\begin{document}

\maketitle

\begin{abstract}
Reproducing research results from papers and released code is central to scientific progress. Existing works have introduced benchmarks to evaluate whether LLM agents can assist with reproducibility, but they are difficult to scale due to their reliance on substantial manual effort for data curation and evaluation.
We introduce \ourmethod, a scalable framework for reproducibility evaluation that leverages human-raised GitHub issues as naturally occurring supervision on realistic reproduction blockers. 
We instantiate \ourmethod on 1,149 recent machine learning papers from major conferences and evaluate four frontier model-agent configurations.
Our results show that LLM agents, even without executing code, can identify many real-world reproducibility problems from paper-repository pairs: the best agent in our study, namely Codex with GPT-5.5, surfaces at least one semantically related human-reported blocker for $\sim 90\%$ of papers in the study. Further analysis shows that agents are particularly effective for surfacing visible failures and identifying the right semantic region, but may still be insufficient in exact localization.
\ourmethod can serve as a reusable, scalable framework for future evaluations of LLM agents on real-world reproducibility auditing. Our code is released at \url{https://github.com/LithiumDA/ReproRepo}.
\end{abstract}

\section{Introduction}
Reproducing previous research is essential for scientific progress, enabling researchers to verify findings and confidently build on prior work~\citep{pineau2021improving,nust2021codecheck}.
As modern research increasingly involves navigating papers, code, data, and other experimental artifacts, there is growing interest in whether large language model (LLM) agents can help automate aspects of reproducibility given their strong coding and reasoning capabilities.
While some recent works challenge agents to produce results from a paper alone~\citep{starace2025paperbench,seo2025paper2code}, the practical utility of reproducibility is in settings where one reproduces results using provided code~\citep{siegel2024corebench,hu2025reprobench}.
This more closely reflects everyday research practice, where researchers build on prior work by rerunning and extending existing codebases and experiments.

A major bottleneck faced by prior work on reproducibility evaluation is that of \emph{scalability}. These approaches require substantial human effort, including expert-curated questions~\citep{siegel2024corebench}, artifact preparation~\citep{ye2025replicationbench}, manually-injected errors ~\citep{shah2026automating}, expert rubrics~\citep{starace2025paperbench,seo2025paper2code}, reference implementation verification \citep{zhao2025autoreproduce}, and manual evaluation~\citep{bai2026story, xu2026scaling}. 
As a result, these datasets typically cover only a limited number of papers---usually fewer than 100---and tend to feature only papers with high-quality artifacts. 
Moreover, when these benchmarks quickly become outdated, the manual work required for update makes it difficult to maintain relevance in the fast-changing research landscape.

\begin{figure*}[t]
\centering
  \includegraphics[width=\linewidth]{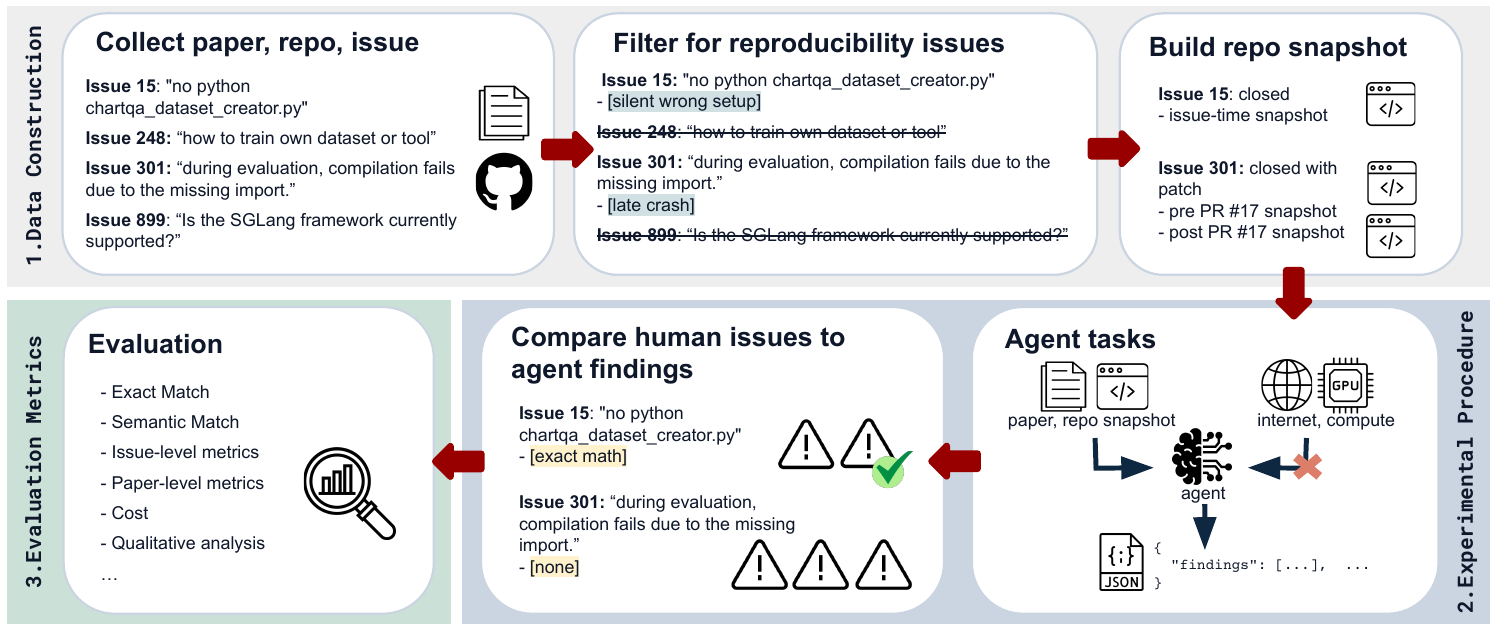}
\caption{\textbf{Overview of \ourmethod}.  
We build an issue-grounded benchmark from paper-repository pairs and human-reported GitHub issues, ask agents to statically audit the paper and fixed repository snapshots, and evaluate whether their findings recover hidden human-reported reproducibility blockers.}
\label{fig:method-overview}
\end{figure*}

\nscomment{Throw some punches here. ReproRepo is based on two key ideas that allow it these properties...you can use bullets to highlight the two key ideas... your current writing has all the content, but you should try to make it a bit more compelling.} \shandacomment{I think the current way shows the flow of our pipeline, i.e., what one would do operationally when using our pipeline (first, then, finally). I'm worried that using bullets may hurt the flow.}

To address this limitation, we introduce \ourmethod, a framework for scalable, realistic, and easily updatable evaluation of reproducibility with LLM agents. 
Our key idea is to \textbf{use real reproduction pain as supervision}: instead of relying on experts to manually design tasks, inject errors, or write evaluation rubrics, \ourmethod leverages GitHub issues as evidence of reproduction blockers encountered by real users.
More specifically, it first builds task instances by collecting conference papers, linked GitHub repositories, and user-reported reproducibility issues.
Then LLM agents are tasked with predicting potential reproducibility blockers given papers and codebases.
Finally, \ourmethod validates agent findings by comparing against human-reported ones and computes match rates as the main evaluation metric. 
In our study, since fully executing modern repositories often requires GPUs, large datasets, and external services such as LLM API calls, we focus on a compute-light static inspection setting for experimentation: agents inspect only the paper and repository snapshot, without code execution.

Using \ourmethod, we evaluate Claude Code and Codex paired with four frontier LLMs on 1149 papers from 3 major machine learning (ML) conferences. 
We find that, surprisingly, static agents (i.e., agents who do not actually execute any code) recover a substantial fraction of human-reported reproduction blockers: Codex with GPT-5.5 identifies at least one semantically related hidden issue for 93.3\% of NeurIPS 2024 main-conference papers and 89.7\% of ICLR 2026 papers. 
Further analysis shows that agents have low false positive rates, are best at surfacing visible failures, and often identify the right semantic region. 
However, they often still struggle with exact localization of errors and identification of less visible failure types.

Overall, the contributions of our work are three-fold:
\nscomment{you already said these things above. Avoid repetition. Merge these to pack punches together with listing  your contributions and findings}\shandacomment{I think this part is supposed to summarize key information above and in some sense repeat the most important things? For readers who just try to skim quickly, they may simply check those sentences without reading the full intro.}
\begin{itemize}
    \item We introduce \textbf{\ourmethod}, a scalable framework for auditing research reproducibility with LLM agents. It uses GitHub issues as naturally occurring human supervision, enabling diverse and continuously updatable benchmarks with minimal manual annotation.
    
    \item We instantiate \ourmethod on recent machine learning papers from major conferences, constructing a large-scale benchmark of 1,149 papers and 7,553 human-reported reproduction blockers.
    
    \item We systematically evaluate frontier LLM agents under \ourmethod. Our results show that static inspection can surface human-reported blockers for most papers, but agents still struggle with precise localization despite strong semantic coverage of visible failures.
\end{itemize}

\begin{table*}[t]
\centering
\resizebox{\linewidth}{!}{%
\begin{tabular}{@{}lcccl@{}}
\toprule
\textbf{Framework} & {\textbf{Given Paper}} & {\textbf{Given Code}} & {\textbf{\# Papers}} & {\textbf{Supervision Source}} \\
\midrule
PaperBench \citep{starace2025paperbench} & \cmark & -- & 20 & Conference papers, expert rubrics \\

Paper2Code \citep{seo2025paper2code} & \cmark & -- & 90 & Conference papers, expert eval \\

AutoReproduce \citep{zhao2025autoreproduce} & \cmark & -- & 13 & Papers w/ verified implementation \\

ReplicationBench \citep{ye2025replicationbench} & \cmark & -- & 31 & Papers, expert tasks \\

Read Paper, Write Code \citep{kohler2026read} & $\sim$ & -- & 48 & Expert-verified reproducible papers \\

\midrule

CORE-Bench \citep{siegel2024corebench} & -- & \cmark & 90 & Papers w/ repos, expert-curated questions \\

R-based social science study \citep{shah2026automating} & -- & \cmark & 5 & Reproducible studies, expert-injected failures \\

\midrule

Scaling Reproducibility \citep{xu2026scaling} & \cmark & \cmark & 384 & Journal papers, expert review \\

REPRO-Bench \citep{hu2025reprobench} & \cmark & \cmark & 112 & Papers w/ expert reproducibility report \\

ReplicatorBench \citep{nguyen2026replicatorbench} & \cmark & $\star$ & 19 & Papers w/ expert reproducibility report \\

\midrule
\textbf{Ours (\ourmethod)} & \cmark & \cmark & \textbf{1149} & Conference papers, GitHub issues \\
\bottomrule
\end{tabular}
}
\caption{\textbf{Comparing \ourmethod with prior reproducibility evaluations.} 
The works are ordered by input setting: paper-only, code-only, and paper-plus-code. 
Checkmarks indicate whether the task includes the listed input; ``$\sim$'' indicates methods description of papers; ``$\star$'' indicates task-dependent code access. 
}
\label{tab:benchmark-comparison}
\end{table*}

\section{Related Work}

\paragraph{LLM agents for scientific workflows}
Before LLM agents, reproducibility evaluation relied on reporting checklists, submission policies, and independent execution or reanalysis \citep{pineau2021improving,magnusson2023reproducibility,stojnic2020codechecklist,nust2021codecheck}.
These efforts clarified review standards but depended on scarce expert labor.
Recent advances of LLM agents have led to a range of works exploring their potential 
in resolving repository issues 
\citep{jimenez2024swebench}, performing research implementations \citep{tian2024scicode,yan2025lmrbench,sun2026co,merrill2026terminalbench,li2026codepde,li2026comind,lyu2026mlsbench}, assisting checklist review \citep{goldberg2024checklist}, verifying scientific manuscripts \citep{liu2023reviewerGPT,son2025spot,xi2025flaws}, assessing research proposal soundness \citep{ho2026soundnessbenchaiscientistreally}, supporting data-driven discovery \citep{chen2024scienceagentbench}, assessing soundness of automated research~\citep{luo2025the} and even rediscovering established findings \citep{wang2026firebench}.
These agent capabilities made it possible and easier to automate reproducibility evaluation workflows.

\paragraph{Agentic reproducibility evaluation.}
Recent works applying LLM agents to reproducibility evaluation can largely be categorized by input type: paper-only, code-only, paper-plus-code.
Paper-only works ask agents to recreate results or implementations from a manuscript or paper-derived description \citep{starace2025paperbench,seo2025paper2code,zhou2025repro,zhao2025autoreproduce,ye2025replicationbench,kohler2026read}.
These tasks often focus more on code generation and measure scientific understanding. Code-only and paper-plus-code works utilize the released code and ask agents to answer curated questions \citep{siegel2024corebench}, repair errors in social science research codebases \citep{shah2026automating}, conduct reanalysis of data \citep{xu2026scaling}, detect fabrications \citep{chen2026fabscore}, score paper-code reproducibility \citep{hu2025reprobench,zhang2026paperrepro}, and evaluate claim replicability \citep{nguyen2026replicatorbench}.
\Cref{tab:benchmark-comparison} summarizes how our work compares with the prior benchmark works. 
In particular, our work focuses on the paper-plus-code setting and leverages GitHub issues for scalability.

\section{\ourmethod}

\nscomment{I read through the section and it is not clear how this section differs from section 4. Section 4 contains some evaluation setup and dataset construction. But this section also contains some sort of evaluation setup and statistics and dataset construction?}

In this section, we describe our proposed framework, \ourmethod, for evaluating LLM agents on reproducibility auditing (Figure~\ref{fig:method-overview}).

\subsection{Data Construction}

\nscomment{how do you know the issues were raised by humans?} \shandacomment{good question. For issues raised before 2025 they are very likely from humans. Since 2025, humans may use coding agent for following others' research but I would trust them to verify an issue before raising the issue. Interested in your take here.} \nscomment{then justify it in the text}\shandacomment{justified}

Our data curation strategy is motivated by how reproducibility problems arise in ordinary research practice. Researchers commonly release code on GitHub, and follow-up users often open issues when they encounter blockers while trying to reproduce the paper. We treat these issues as user-reported records of reproduction experiences: even if some recent reports are drafted with AI assistance, opening an issue still reflects a user's judgment that the blocker is real and relevant. These issues therefore provide naturally occurring records of reproduction experiences and can serve as free supervision signals.
Moreover, GitHub issues come with useful structure, such as timestamps, open/closed status, discussion threads, and, in some cases, linked commits or pull requests that document how a problem was resolved. 
Our data construction pipeline is built upon these properties. 

\paragraph{Data collection \& filtering.}
We construct evaluation data from conference papers, their associated GitHub repositories, and user-reported GitHub issues. 
Starting from a complete set of accepted or published conference papers, we collect paper metadata and GitHub repository URLs from Paper Copilot \citep{yang2026paper} and conference metadata. Non-GitHub links and papers without a usable repository are filtered out. 
For each repository, we collect both open and closed GitHub issues. Open issues provide real user reports. Closed issues provide historical reports that may have been clarified or fixed after the original failure. Then an LLM judge is employed to filter for reproducibility-related issues.



\paragraph{Repository snapshot construction.}
To prevent information leakage and post hoc reasoning during reproducibility evaluation, we build snapshots differently for open and closed issues.
For open issues, we collect the repository's default-branch state at collection time as the codebase snapshot. 
For closed issues, we collect the issue-time codebase snapshot, i.e., the repository state immediately before the issue creation, to make sure the evaluation presents to agents a pre-fix code state that a user would have seen when reporting the problem. 
For closed issues with patch evidence, we additionally recover a post-fix snapshot from the human resolution. This will be used as an additional validation set, i.e., a target problem should disappear after the human fix, to evaluate the false positive rate of predicted issues from agents.


\subsection{Reproducibility Audit Protocol}
Given the constructed paper-repository snapshots and hidden human issues, we next describe how agents perform static audits and how their outputs are assessed.

\paragraph{Agent tasks.}
We task LLM agents with identifying potential reproducibility blockers given papers and codebases as input. 
The curated human-reported issues are hidden from the agents.
\nscomment{how? could the agent fetch it from the internet? is there a concern of data contamination?}  \shandacomment{We already mentioned below: \textit{(we) deny the agent internet access.}}\nscomment{contamination doesnt happen just because of internet access. it can happen because the data was in its pretraining data}\shandacomment{in \Cref{sec:exp-main-result} we find the results remain stable across venues and years, so agents are not likely to solve the task by memorization}
Agents are expected to output a list of predicted reproduction blockers, ordered by the estimated likelihood of the blocker encountered by real users.
Each finding should describe one concrete, GitHub-issue-style problem that a real reproducer would plausibly report.
For each run, we prepare an isolated workspace that contains the code repository snapshot and the corresponding paper file. To prevent information leakage and post hoc reasoning, we remove all git commit histories and deny the agent internet access. 
Our evaluation particularly considers a compute-light static inspection setting. The agent is instructed to perform a static inspection without environment setup and code execution. GPU compute is withheld from the workspace. The full prompt is provided in \Cref{app:discovery-prompt}.


\paragraph{Agent output assessment.}
After prediction, we evaluate the agent output with a separate LLM judge. The judge receives the hidden human issues for the same paper-snapshot pair as well as the agent's blind findings. For each hidden human issue, the judge compares it against the agent findings and assigns a label among \textit{exact match} (EM), \textit{semantic match} (SM), and \textit{none} based on the following criteria: 
\nscomment{an example will be very helpful. a figure may be an overkill}\shandacomment{Yeah we have been manully looked into the cases and I have some candidate. Will add some in the appendix.}\nscomment{if it is relatively short, add it here}\shandacomment{in appendix}

\begin{itemize}[leftmargin=*,style=nextline]
  \item \textbf{EM}: The finding identifies substantially the same practical reproduction problem and the same trigger, even if the wording might differ.
  \item \textbf{SM}: The finding identifies a semantically related blocker with concrete overlap yet incomplete specificity, e.g., the same evaluation pipeline but different failure points, the same missing/released artifact family but different files or checkpoints, etc.
  \item \textbf{None}: No agent finding would be useful evidence that the agent discovered this issue.
\end{itemize}
Concrete examples of the three labels, including verbatim agent-output excerpts, are provided in \Cref{app:alignment-examples}.
The full judge prompt is provided in \Cref{app:alignment-judge-prompt}.

\begin{table*}[t]
\centering
\resizebox{\linewidth}{!}{%
\begin{tabular}{@{}lccccc@{}}
\toprule
Conference & \# accepted papers & \# selected papers & \# open issues & \# closed issues & \# closed issues w/ patch \\
\midrule
NeurIPS 2022 main & 2905 & 264 & 579 & 974 & 22 \\
NeurIPS 2022 DB & 163 & 88 & 342 & 524 & 32 \\
NeurIPS 2024 main & 4037 &  410 & 1642 & 1518 & 40 \\
NeurIPS 2024 DB &  460 & 107 & 299 & 678 & 44 \\
ICLR 2026 & 5355 & 280 & 405 & 592 & 12 \\
\midrule
All datasets & 12686 & 1149 & 3267 & 4286 & 150 \\
\bottomrule
\end{tabular}
}
\vspace{-2mm}
\caption{\textbf{Dataset Statistics.} Accepted-paper counts are conference- or track-level public counts. Selected papers are papers retained after issue filtering, i.e., papers with at least one accepted benchmark issue. Open and closed issue counts refer to accepted benchmark cases. DB denotes datasets and benchmarks track.}
\label{tab:data-overview}
\end{table*}

\subsection{Evaluation Metrics}
\label{sec:method-metrics}
We evaluate agent predictions based on the assessment outcome described above. The main metrics are EM rate and SM rate, detailed below.



\paragraph{Top-$k$ reporting budget.}
The agent outputs an ordered list of findings for each paper/repository snapshot. To make scores comparable across agents and to avoid rewarding arbitrarily long reports, we evaluate matches under a fixed reporting budget. For a budget $k$, only the top $k$ findings are eligible to match hidden issues. We write EM@$k$ and SM@$k$ for exact and semantic matching under this budget. We use $k=10$ in the main experimental results. We also report curves over $k$ to show how performance changes as the reviewer-facing budget increases in \Cref{sec:exp-ablation}.

\paragraph{Result aggregation level.} Exact match and semantic match can be aggregated across different granularities. Let $F_p^{(k)}$ denote the top $k$ agent findings for paper or snapshot group $p$. A hidden issue $i$ associated with $p$ is counted as matched at $k$ only if its best-aligned finding lies in $F_p^{(k)}$.

The \textit{issue-level metric} treats each hidden issue as one evaluation item. Let $I$ be the set of hidden issues, and let $y_i^{(k)}$ be the best label for issue $i$ under the top-$k$ budget. We define:
\[
\begin{aligned}
\mathrm{Issue}_{\mathrm{EM}}@k
  &= \frac{|\{i \in I : y_i^{(k)}=\mathrm{EM}\}|}{|I|}; \\
\mathrm{Issue}_{\mathrm{SM}}@k
  &= \frac{|\{i \in I : y_i^{(k)}\in\{\mathrm{EM},\mathrm{SM}\}\}|}{|I|}.
\end{aligned}
\]

Issue-level match rate measures how much of the hidden issue evidence is recovered by the agent. It is the most direct issue-retrieval metric, but it can be strict for papers with many related issues, since each issue is counted separately.

The \textit{paper-level metric} characterizes whether the static audit surfaces one or all human-observed reproduction blockers for a paper. Let $P$ be the set of evaluated papers, and let $I_p$ be the set of hidden issues associated with paper $p$. The \emph{any-issue} paper match rate counts a paper as match if at least one of its hidden issues is matched, while the \emph{all-issue} paper match rate counts a paper only if all the hidden issues are matched.
\[
\begin{aligned}
\mathrm{Paper}^{\mathrm{any}}_{\mathrm{EM}}@k  =
\frac{\left|\left\{p \in P :
\substack{\exists~i \in I_p \text{ s.t.}\\ i \text{ is EM@} k}
\right\}\right|}{|P|};\\
\mathrm{Paper}^{\mathrm{all}}_{\mathrm{EM}}@k =
\frac{\left|\left\{p \in P :
\substack{\forall~i \in I_p,\\ i \text{ is EM@} k}
\right\}\right|}{|P|}.
\end{aligned}
\]

$\mathrm{Paper}^{\mathrm{any}}_{\mathrm{SM}}@k$ and $\mathrm{Paper}^{\mathrm{all}}_{\mathrm{SM}}@k$ are defined similarly. 
This paper-level coverage metric is practically relevant. For example, identifying one real blocker can already direct reviewer or author attention to a problematic artifact in paper triage.

\paragraph{False positive rate.}
High match rates are useful only if agent findings are not dominated by generic or persistent warnings that remain even after a real problem has been fixed.
We therefore evaluate false positives using closed issues with clear code patches.
For each such issue, we run the agent on the \textit{post-fix} repository snapshot, where the corresponding historical blocker should no longer be present.
If the agent still reports a finding that matches the fixed issue, we count it as a false positive.
In \Cref{sec:exp}, we aggregate FPR at the paper level: a paper is counted as false positive if any predicted finding matches an already-fixed issue.
This is a conservative criterion and provides an upper bound on issue-level false positives.


\begin{figure*}[t]
\centering
\includegraphics[width=0.287\linewidth]{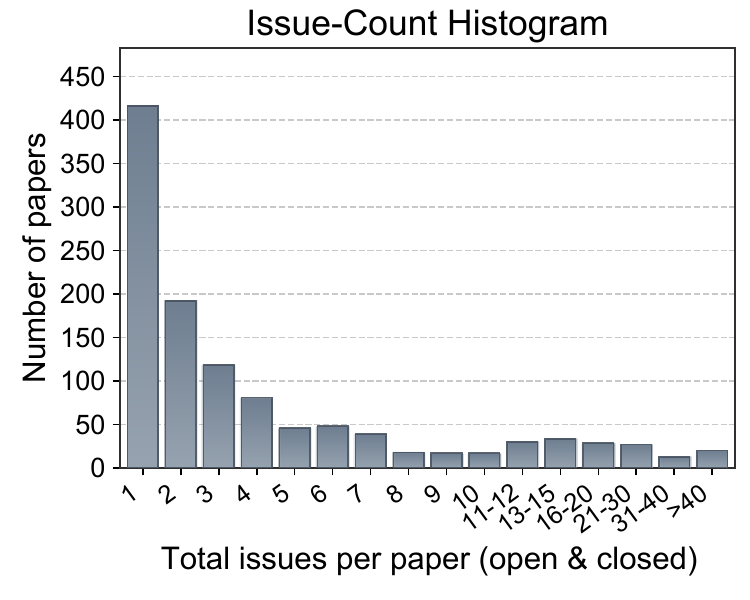}
\includegraphics[width=0.287\linewidth]{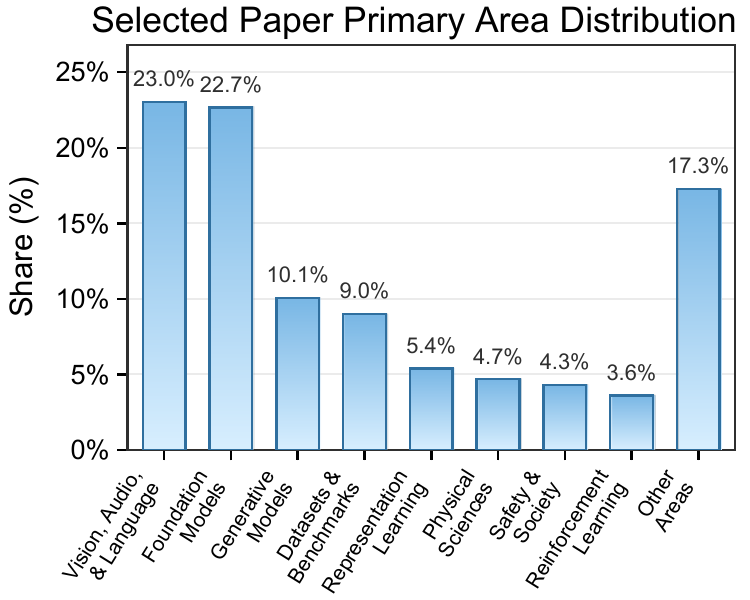}
\includegraphics[width=0.41\linewidth]{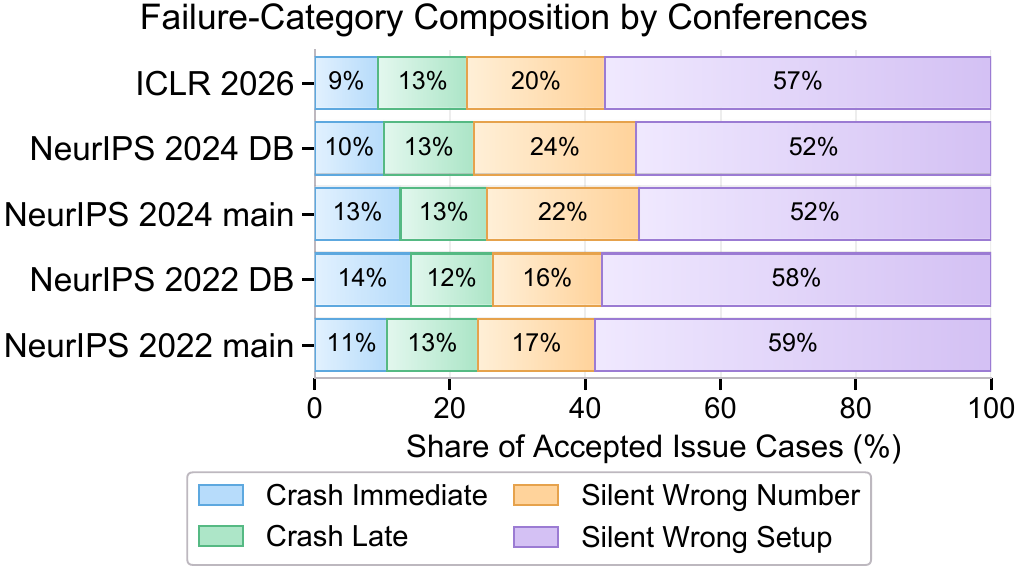}
\vspace{-7mm}
\caption{\textbf{Dissections on the 1149 selected papers and 7553 issues.} \textbf{Left}: distribution of selected issue counts per paper, aggregated across all five conferences. \textbf{Middle}: primary research area distribution of selected ICLR 2026 papers where area metadata is available. \textbf{Right}: composition of failure categories in the selected issues. DB denotes datasets and benchmarks track.}
\label{fig:dataset-dissection}
\end{figure*}

\begin{table*}[!t]
\centering
\resizebox{\linewidth}{!}{%
\begin{tabular}{@{}llcccccccc@{}}
\toprule
Model & Agent
& \multicolumn{2}{c}{Issue Match}
& \multicolumn{2}{c}{Paper Match (Any)}
& \multicolumn{2}{c}{Paper Match (All)}
& FPR
& Cost \\
\cmidrule(lr){3-4}\cmidrule(lr){5-6}\cmidrule(lr){7-8}
& & EM@10 & SM@10
& EM@10 & SM@10
& EM@10 & SM@10
& & \\
\midrule
DeepSeek-V4-Pro & Claude Code
& 19.3\% & 53.1\%
& 47.6\% & 89.0\%
& 25.6\% & 55.3\%
& 2.80\% & \$0.29 \\
Claude Opus 4.7 & Claude Code
& 21.7\% & 54.1\%
& 52.0\% & 86.5\%
& 27.8\% & 57.1\%
& 2.10\% & \$4.07 \\
GPT-5.4-Mini & Codex
& 13.4\% & 43.4\%
& 33.0\% & 82.1\%
& 18.3\% & 52.0\%
& \textbf{1.40\%} & \$0.22 \\
GPT-5.5 & Codex
& \textbf{25.4\%} & \textbf{58.1\%}
& \textbf{59.7\%} & \textbf{89.7\%}
& \textbf{34.8\%} & \textbf{65.6\%}
& 3.50\% & \$1.26 \\
\bottomrule
\end{tabular}
}
\vspace{-0mm}
\caption{\textbf{Comparing Model Performances.} EM@10 and SM@10 denote exact match rate and semantic match rate among the first 10 agent findings. All match metrics are computed on the ICLR 2026 paper set. FPR is computed on the full post-fix validation subset across all five conferences/tracks.
Issue match is computed over hidden issues. Paper Match (Any) counts a unique paper as match if at least one hidden issue for that paper is matched, while Paper Match (All) requires all hidden issues for that paper to be matched. 
FPR denotes false positive rate on the post-fix snapshots of all closed issues with patches.
Cost denotes the average cost per run.
}
\label{tab:model-comparison-10}
\end{table*}

\section{Evaluation Setup}
\subsection{Data Overview}
\label{sec:data}
In our study, we focus on machine learning conferences because code release is common in this community and paper metadata is readily available from \href{https://openreview.net}{OpenReview}. 
To cover artifacts from both earlier and recent venues, we include NeurIPS 2022 and NeurIPS 2024, each separated into the main conference and the datasets \& benchmarks track, as well as ICLR 2026.
Starting from accepted papers with public GitHub repository links, we collect candidate open and closed GitHub issues and retain papers with at least one reproducibility-related issue. 
This collection and filtering procedure is venue-agnostic and can be readily extended to additional conferences or journals.

\paragraph{Dataset statistics.}
\Cref{tab:data-overview} provides an overview of the data scale in our evaluation. 1149 papers and 7553 human-reported reproducibility issues are selected from a set of 12686 papers, making the evaluation significantly larger than all the prior studies. In particular, there are 150 closed issues with clear code patches that address the issue, which will be used to examine false positive rates of the agent prediction.

\paragraph{Dataset dissection.}
\Cref{fig:dataset-dissection} summarizes three views of the resulting benchmark.
The left panel shows that the number of selected issues per paper follows a long-tail distribution: most papers contribute only a handful of reproducibility issues, while a small set of heavily-used repositories contributes many. Per-conference histograms are deferred to \Cref{fig:issue_count_histograms}.
The middle panel shows that selected papers span a diverse range of primary research areas, with Vision, Audio \& Language Applications (23.0\%) and Foundation Models (22.7\%) as the largest groups, followed by a long tail of other areas.
The right panel shows that the four failure categories, namely crash immediate, crash late, silent wrong setup, and silent wrong number, detailed in \Cref{app:failure-taxonomy}, appear in stable proportions across all five conferences/tracks, with \emph{silent wrong setup} as the dominant category (52--59\%). 
The results suggest that \ourmethod captures a general distribution of diverse real-world reproducibility blockers.

\subsection{Models \& Agents}
Our evaluation covers two popular coding agents, Claude Code and Codex, paired with four frontier open and closed models: DeepSeek-V4-Pro~\citep{deepseekai2026deepseekv4}, Claude Opus 4.7~\citep{anthropic2026claudeopus47}, GPT-5.4-Mini~\citep{openai2026gpt54mini}, and GPT-5.5~\citep{openai2026gpt55}. DeepSeek-V4-Pro and Claude Opus 4.7 are evaluated with Claude Code, and GPT-5.4-Mini and GPT-5.5 with Codex. We use GPT-5.5 as the LLM judge.

\begin{table*}[t]
\centering
\resizebox{\linewidth}{!}{%
\begin{tabular}{lcccccccc}
\toprule
Conference
& \multicolumn{2}{c}{Issue Match}
& \multicolumn{2}{c}{Paper Match (Any)}
& \multicolumn{2}{c}{Paper Match (All)}
& Token Usage
& Cost \\
\cmidrule(lr){2-3}\cmidrule(lr){4-5}\cmidrule(lr){6-7}
& EM@10 & SM@10
& EM@10 & SM@10
& EM@10 & SM@10
& & \\
\midrule
NeurIPS 2022 main
& 23.8\% & 57.4\%
& 61.2\% & 89.7\%
& 15.2\% & 41.4\%
& 966,794
& \$1.28 \\
NeurIPS 2022 DB
& 18.2\% & 51.0\%
& 60.2\% & 96.6\%
& 10.2\% & 31.8\%
& 1,225,375
& \$1.50 \\
NeurIPS 2024 main
& 23.9\% & 56.3\%
& 61.0\% & 93.3\%
& 22.7\% & 51.1\%
& 925,068
& \$1.30 \\
NeurIPS 2024 DB
& 12.2\% & 47.3\%
& 46.7\% & 87.6\%
& 10.5\% & 33.3\%
& 1,159,698
& \$1.53 \\
ICLR 2026
& 25.4\% & 58.1\%
& 59.7\% & 89.7\%
& 34.8\% & 65.6\%
& 943,011
& \$1.26 \\
\bottomrule
\end{tabular}
}
\caption{\textbf{Comparing Results on Different Conferences.} EM@10 and SM@10 denote exact match rate and semantic match rate among the first 10 agent findings, and the results are based on Codex with GPT-5.5.
Issue match is computed over hidden issues. Paper Match (Any) counts a unique paper as match if at least one hidden issue for that paper is matched, while Paper Match (All) requires all hidden issues for that paper to be matched. 
Token usages refers to the average number of total tokens per run, including both input and output.
Cost denotes the average cost per run.}
\label{tab:conference-comparison-10}
\end{table*}

\begin{figure*}[t]
\centering
\includegraphics[width=0.495\textwidth]{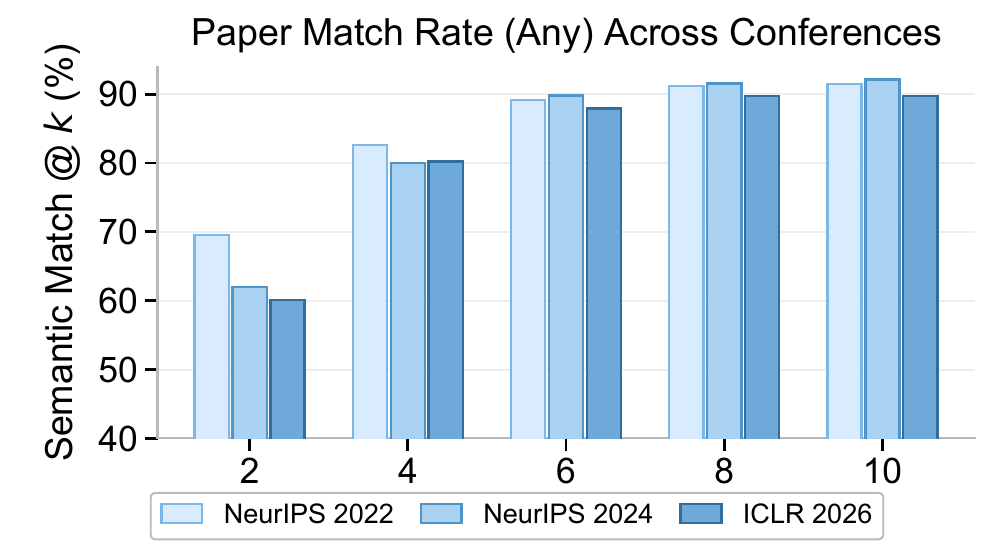}
\includegraphics[width=0.495\textwidth]{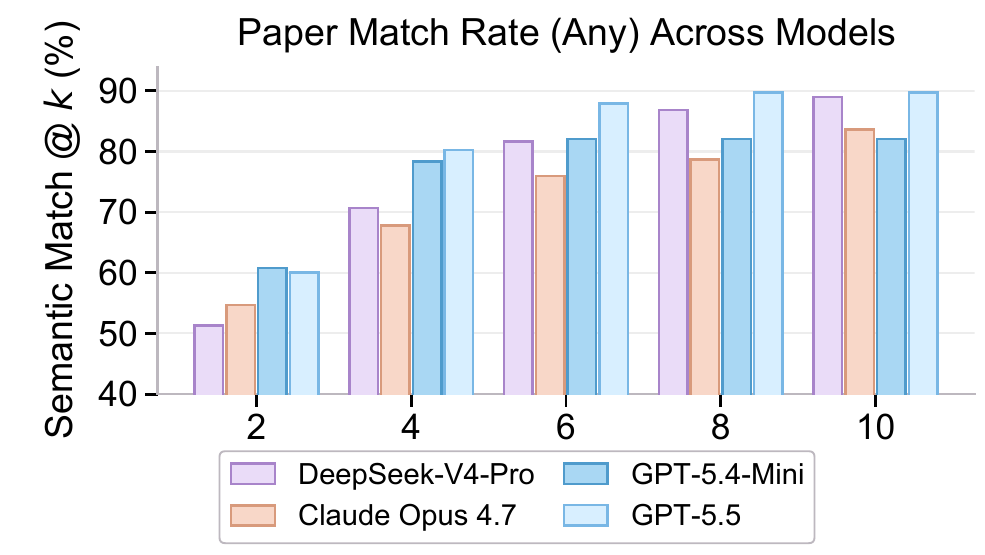}
\vspace*{-0.2cm}
\caption{\textbf{Semantic match rates with varying reporting budget $k$}. \textbf{Left}: conference comparison for GPT 5.5 with Codex, with the two NeurIPS tracks combined by year. \textbf{Right}: model comparison on ICLR 2026.}
\label{fig:topk-paper-match}
\end{figure*}

\section{Experimental Results}
\label{sec:exp}

In this section, we present our experimental results and discuss the key findings.

\subsection{Main Evaluation Result}
\label{sec:exp-main-result}

\nscomment{please keep each table and figure near to where they are referred. table 3 is 3 pages ago.}\shandacomment{isn't table 3 at the top of this pg?}
\nscomment{why is iclr 2026 themain result as opposed to also having neurips?}\shandacomment{just bc it is the most recent one}
\Cref{tab:model-comparison-10} compares the four models in our study on the ICLR 2026 paper set. \Cref{tab:conference-comparison-10} further reports the performance of Codex with GPT-5.5, the best-performing configuration in \Cref{tab:model-comparison-10}, across five conferences/tracks. We make the following observations based on the results:

\textbf{Surprisingly, agents can consistently identify reproducibility blockers through static inspection.}
\Cref{tab:model-comparison-10} shows that all four models recover at least one human-reported issue for a large fraction of papers without executing the code. Under semantic matching, the paper-level any-issue match rate ranges from 82.1\% to 89.7\%; even under the stricter exact match criterion, the state-of-the-art GPT-5.5 model achieves a rate of 59.7\%. \Cref{tab:conference-comparison-10} further shows that this trend holds across all five conference/track splits. These results suggest that many real reproducibility blockers leave visible traces in the paper-repository pair. Meanwhile, we note that while DeepSeek-V4-Pro is less accurate under exact matching, it reaches a similar paper-level semantic match rate to GPT-5.5 (89.0\% vs. 89.7\%) while costing only 23\% as much per run. This makes it a strong cost-quality trade-off when broad triage coverage is the main goal.

\textbf{Agents often identify the right semantic region, but exact localization remains challenging.}
Across both \Cref{tab:model-comparison-10} and \Cref{tab:conference-comparison-10}, semantic match rates are substantially higher than exact match rates. For example, in the model comparison, issue-level SM@10 ranges from 43.4\% to 58.1\%, while issue-level EM@10 ranges only from 13.4\% to 25.4\%. A similar gap appears at the paper level. This indicates that agents can often identify the relevant workflow, artifact, or failure region, but do not always recover the precise trigger or root cause reported by users. Therefore, semantic matches should be interpreted as useful triage signals rather than confirmed localization of exact issues.

\textbf{The results remain stable across venues and years.}
\Cref{tab:conference-comparison-10} shows no clear evidence that the agent performs better on older or later venues. The issue-level semantic match rates are close for NeurIPS 2022 main, NeurIPS 2024 main, and ICLR 2026, and the paper-level any-issue semantic match rate remains around 90\%. This stability suggests that the task is unlikely to be solved mainly by memorization.\footnote{The knowledge cutoffs are May 2025 for DeepSeek V4, December 2025 for GPT-5.5 and January 2026 for Claude Opus 4.7. Most ICLR 2026 repositories in our benchmark were not publicly released until after models' knowledge cutoffs.} 
One notable trend is that Datasets and Benchmarks tracks tend to have lower issue-level and all-issue match rates than the corresponding main conferences, indicating that their reproduction blockers may be more subtle or harder to identify. They also require more tokens on average, consistent with the larger or more complex artifact structure in these papers.

\nscomment{there are a number of interesting nuggets in section 5. However, most readers wont see this. They will see the intro. Identify the most interesting/important findings and put a bulleted list of them in the intro.}\shandacomment{the first two findings have been mentioned in the abs bc they are the most relevant.} 

\textbf{The false positive rate is very low.}
In \Cref{tab:model-comparison-10}, the false positive rate is at most 3.50\% across models, much lower compared with paper-level match rates. This is encouraging because the false-positive evaluation is conducted on post-fix snapshots, where the matched historical issue should no longer be present. The result indicates that the agents' findings are not dominated by hallucinated blockers and that most reported issues correspond to plausible static evidence.

\subsection{Ablation Study}
\label{sec:exp-ablation}

\begin{figure}[t]
\centering
\includegraphics[width=\linewidth]{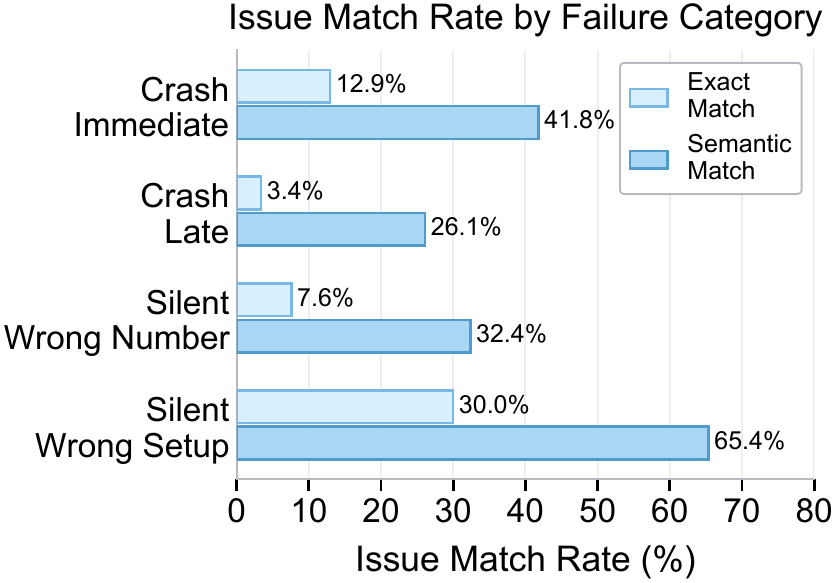}
\caption{\textbf{Exact and semantic match rates} of GPT-5.5 on different failure categories.}
\label{fig:issue-category}
\end{figure}

\paragraph{Effect of the reporting budget.}
A natural concern is that the paper-level match rate may be driven mainly by allowing the agent to report many candidate findings.
To address this concern, we plot the match rate under top-$k$ reporting budget for varying $k$ in \Cref{fig:topk-paper-match}.
The result shows that the match rates always rise quickly at small $k$. 
Across conferences, semantic paper-level match is already above 60\% at $k=2$ and around 80\% or higher at $k=4$, before plateauing at roughly 90\% by $k=6$ to $10$. 
The model comparison shows the same pattern: most of the gain appears in the first few ranked findings, while later findings provide smaller
incremental coverage. 
This suggests that the high match rate does not come from long issue lists. 
Our ranked static audit requirement encourages the model to place real reproduction blockers near the top of the report.
The property is practically useful for reviewer or author triage when humans only have limited bandwidth to check a few potential issues.

\paragraph{Ablation on paper input.} 
One important aspect of reproducibility is paper-code consistency. 
Simply inspecting codebases without access to the paper may not be sufficient for identifying all reproduction blockers, since many failures are defined only relative to the paper's claimed datasets, checkpoints, evaluation metrics, etc. 
We validate this hypothesis by comparing a code-only setting with the default paper-and-code setting in \Cref{tab:code_only_ablation}. 
Adding the paper improves performance across all metrics, with the largest gains appearing under exact matching: EM@10 increases by 8.98 percentage points at the paper-level any-issue metric. 
This suggests that repository inspection alone can often identify the broad failure region, as reflected by the strong code-only semantic match rates, but access to the paper helps the agent determine whether a repository artifact, command, configuration, or released resource actually supports the paper's intended reproduction target. 
Therefore, paper input is most valuable for turning general repository-risk detection into more precise paper-grounded reproducibility auditing.

\subsection{Issue Taxonomy}
\label{sec:exp-issue-taxonomy}

\begin{table}[t]
\centering
\resizebox{\linewidth}{!}{%
\begin{tabular}{lr}
\toprule
Finding type & Count \\
\midrule
Documentation-repository mismatch & 27 \\
Dependency or environment gap & 23 \\
Missing or under-specified artifacts & 23 \\
Missing evaluation/reproduction workflow & 13 \\
Paper-code protocol mismatch & 3 \\
\bottomrule
\end{tabular}
}
\vspace{-1mm}
\caption{\textbf{Manual categorization} of 89 top-ranked unmatched \codex findings from papers with one collected human issue.}
\label{tab:unmatched-validity}
\end{table}

\begin{table}[t]
\centering
\resizebox{\linewidth}{!}{%
\begin{tabular}{@{}lcccc@{}}
\toprule
Agent input
& \multicolumn{2}{c}{Issue Match}
& \multicolumn{2}{c}{Paper Match (Any)} \\
\cmidrule(lr){2-3}\cmidrule(lr){4-5}
& EM@10 & SM@10
& EM@10 & SM@10 \\
\midrule
Code only
& 20.9\% & 57.0\%
& 50.7\% & 87.6\% \\
Paper \& code
& \textbf{25.4\%} & \textbf{58.1\%}
& \textbf{59.7\%} & \textbf{89.7\%} \\
\bottomrule
\end{tabular}
}
\vspace{-1.5mm}
\caption{\textbf{Ablation on agent input.} Experiment is conducted on ICLR 2026 paper set with GPT-5.5.}
\label{tab:code_only_ablation}
\end{table}

We next examine which kinds of reproducibility issues appear in the benchmark and which of them are visible to static auditing.
\Cref{fig:dataset-dissection} (right) summarizes the distribution of the four human issue categories, and \Cref{fig:issue-category} reports category-level match rate.
The category definitions are given in \Cref{app:failure-taxonomy}.
We defer the corresponding agent-prediction category distributions to \Cref{fig:agent-prediction-category-all,fig:agent-prediction-category-exact-match,fig:agent-prediction-category-semantic-match}.
We make two observations:

\textbf{The issue taxonomy is consistent across conferences.}
\Cref{fig:dataset-dissection} (right) shows that the 5 conferences/tracks share highly similar patterns in their failure-category composition.
Specifically, \emph{Silent Wrong Setup} is consistently the largest failure category. It covers more than half of the issues, including mismatch of the paper's intended configuration, inconsistency in the artifact, etc.
This suggests that the benchmark is not driven by a venue-specific artifact norm; across datasets, many reproduction blockers arise before long training runs or final metric comparison.

\textbf{Static auditing works better for visible repository failures.}
\Cref{fig:issue-category} shows that the agent recovers \emph{silent wrong setup} most reliably, followed by \emph{crash immediate}.
This pattern is consistent with static inspection: missing checkpoints, undocumented data preparation, broken imports, mismatched scripts, absent evaluation commands, and paper-repository scope gaps often leave evidence in READMEs, configs, file structure, or paper claims.
By contrast, \emph{crash late} and \emph{silent wrong number} are harder to recover statically because they often require execution, downloaded artifacts, hardware-specific paths, long-horizon training, or comparison against reported metrics.
Taken together, these results suggest that no-execution auditing is useful for early triage of visible setup, artifact, and evaluation-path risks, but cannot replace execution or output comparison.

\subsection{Additional Analyses}

We further examine whether unmatched top-ranked findings are unsupported predictions or parallel reproducibility concerns not reported by humans.
For this analysis, we focus on 89 ICLR 2026 papers with exactly one human-reported issue, where the top-ranked \codex finding does not match that issue.
We manually categorize each finding using static repository evidence, including documentation, scripts, configs, dependency files, file inventory, and paper-claim alignment.

Manual review shows that \textbf{most unmatched findings reflect parallel reproducibility risks.}
All 89 findings had concrete static evidence for a plausible issue in the released artifact.
The type breakdown is presented in Table \ref{tab:unmatched-validity}.
These findings can fail to match the single collected human issue for several reasons: a user may report only the first blocker encountered, some gaps can be repaired locally without filing an issue, and paper-level reproduction gaps such as missing aggregation scripts may not appear during ordinary package use.
Thus, in this low-issue-count subset, non-matching top-ranked agent findings more often reflect sparse human supervision and parallel reproduction risks than unsupported predictions.

\section{Conclusion}
We introduce \ourmethod, a scalable framework for evaluating LLM agents on issue-grounded static reproducibility auditing. 
By using human-reported GitHub issues as naturally occurring supervision, \ourmethod enables large, realistic, and easily updatable evaluation over paper-repository pairs. 
Across 1,149 machine learning papers and 7,553 reproducibility blockers curated from recent major conferences, we find that static agents can surface many real user-facing failures without executing code, identifying at least one semantically related issue for roughly 90\% of papers with the best configuration, i.e., Codex with GPT-5.5. 
However, currently, these agents still struggle with the exact localization of the blockers. 
In the future, \ourmethod can serve as a reusable pipeline for evaluating new models' and agents' capabilities to audit the reproducibility of newly released papers.


\section*{Limitations}
\ourmethod is based on GitHub issues. While those human-reported issues provide  evidence from real users and make the benchmark substantially easier to scale, they are noisier than expert-curated labels. Users may report only the first blocker they encounter, some valid reproduction risks may never be filed as issues, and a small number of issues may come from user mistakes or reflect user-specific environments or transient external states. This limitation can be mitigated through data filtering. 

Another limitation is that the issues do not necessarily exhaust all reproducibility blockers, especially for repositories with few users. Besides, the study mainly focuses on repositories with at least one accepted reproducibility issue, its metrics characterize issue discovery and localization under incomplete issue supervision, rather than repository-level discrimination between problem-free and problematic repositories.

Our experiments focus on static, no-execution auditing. 
This setting is intentionally compute-light and useful for early triage. 
However, it does not cover all reproducibility failures, such as long-running crashes, hardware-specific errors, etc. 
Future work can combine issue-grounded static auditing with execution-based validation for a more complete view of reproducibility.

\section*{Ethical Considerations}

This study uses publicly available repository metadata and GitHub issue discussions as evidence of reproducibility barriers. Our analysis does not require identifying individual issue authors, and any released data should avoid unnecessary personal information such as user handles. 
Automated audit findings should be treated as triage signals rather than definitive judgments about a paper or repository. To reduce the risk of over-interpretation, we evaluate false positives and ground findings in repository evidence. Repository maintainers should have opportunities to correct false positives, clarify documentation, and update artifacts.

\section*{Acknowledgments}
The authors would like to thank Jingchu Gai, Baihe Huang, Yixin Huang, Haowei Lin, Honghao Lin, Naveen Raman, Maitreyi Swaroop, Zhiwei Steven Wu for discussions.

NBS was supported in part by grants NSF 1942124 and ONR N000142512346. TD was supported in part by Schmidt Sciences. AT was supported in part by gift funding from Datadog.

\bibliography{references}

\appendix

\section{Artifact Use, Licenses, \& Intended Use}
\label{app:artifact-use}

Our study builds on existing public artifacts, including conference paper metadata, public GitHub repositories, GitHub issue threads, and repository links discovered from Paper Copilot and conference metadata.
We use these artifacts only for research purposes: constructing an issue-grounded benchmark for evaluating static reproducibility auditing, measuring whether LLM agents can identify human-reported reproduction blockers, and supporting aggregate analyses of reproducibility failures.

Paper Copilot \citep{yang2026paper} is used as a discovery source for paper metadata and repository links, rather than as a source of benchmark labels.
We cite Paper Copilot and do not claim ownership over its content.
Our released artifacts will not relicense or redistribute Paper Copilot content beyond the minimal metadata needed to identify papers and repositories, subject to the applicable terms of use.

GitHub repositories and issues remain governed by their original repository licenses, GitHub terms, and any other applicable upstream conditions.
Our work does not relicense repository code, issue text, papers, or third-party metadata.
When releasing artifacts, we will release only derived annotations, evaluation scripts, paper/repository identifiers, issue identifiers, commit hashes, and instructions for reconstructing snapshots where permitted.
Users of the released artifacts are responsible for complying with the original licenses and access conditions of the corresponding papers, repositories, GitHub content, and metadata sources.

The intended use of the artifacts created by our work is research on reproducibility auditing, benchmark construction, and evaluation of LLM agents.
The artifacts are not intended for ranking, shaming, or making definitive claims about individual authors, maintainers, papers, or repositories.
Agent-generated audit findings should be treated as triage signals that require human verification.
Use of derived artifacts outside research contexts, or in ways that conflict with the original access conditions of the underlying sources, is not intended.


\section{Dataset Details}
\label{app:dataset-details}

\subsection{Per-Conference Issue Counts}
\label{app:issue-counts}

\begin{figure*}[t]
\centering
\begin{minipage}{0.325\textwidth}
  \centering
  \includegraphics[width=\linewidth]{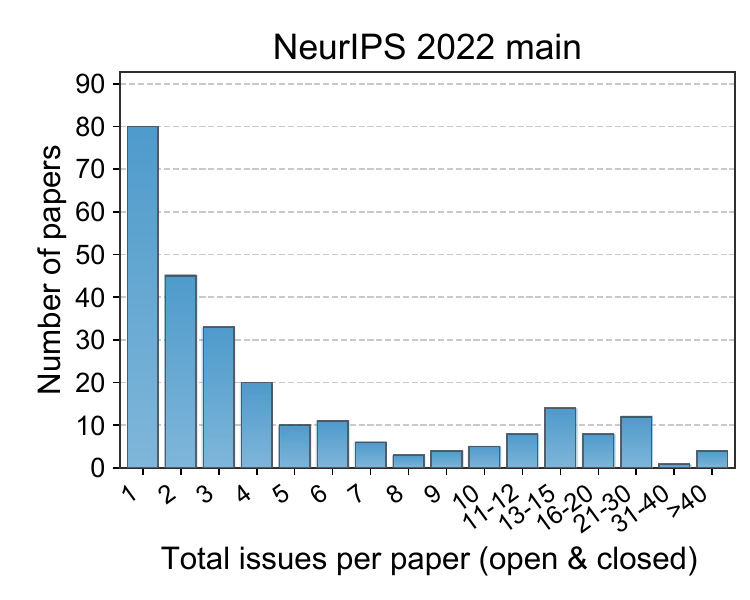}
\end{minipage}
\hfill
\begin{minipage}{0.325\textwidth}
  \centering
  \includegraphics[width=\linewidth]{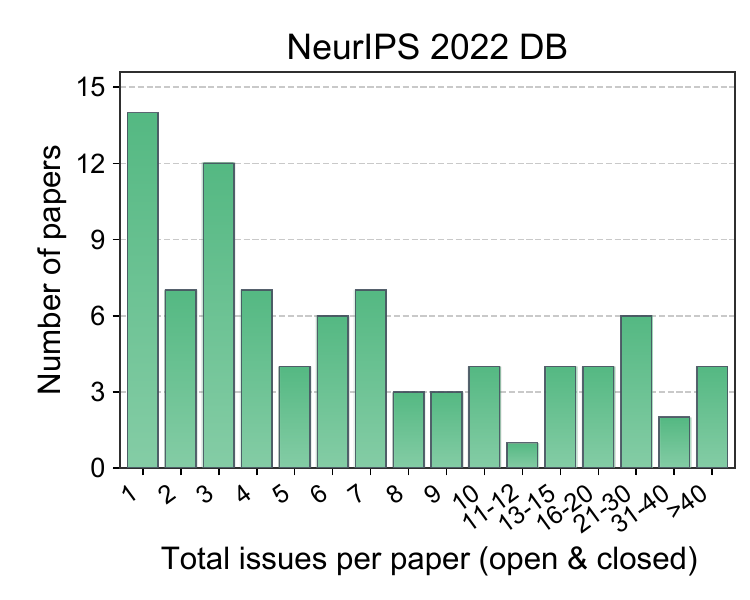}
\end{minipage}
\hfill
\begin{minipage}{0.325\textwidth}
  \centering
  \includegraphics[width=\linewidth]{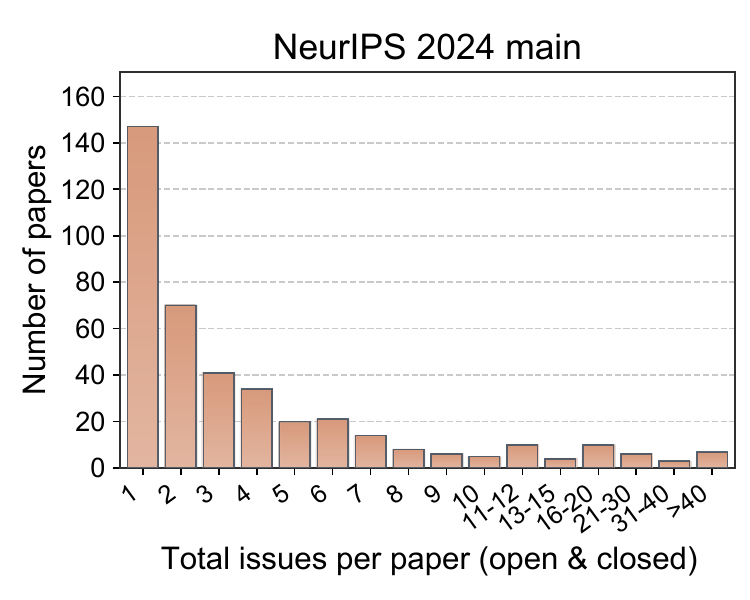}
\end{minipage}

\begin{minipage}{0.325\textwidth}
  \centering
  \includegraphics[width=\linewidth]{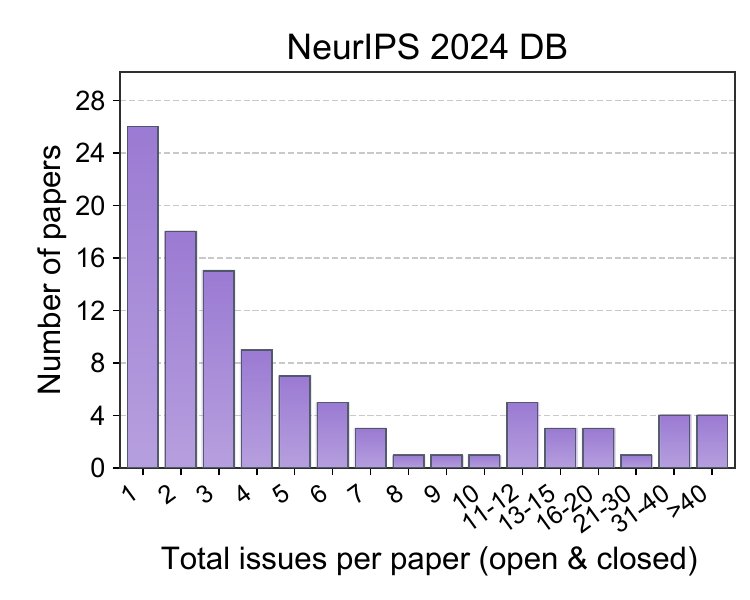}
\end{minipage}
\begin{minipage}{0.325\textwidth}
  \centering
  \includegraphics[width=\linewidth]{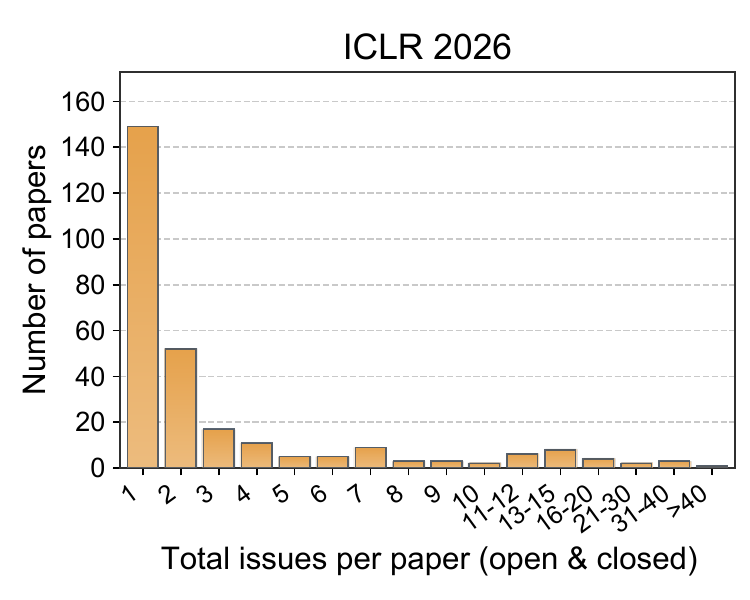}
\end{minipage}
\caption{\textbf{Issue-Count Skew.} Per-split frequency distributions of hidden
issue counts per paper, computed as the sum of open and closed reproduction
issues for each issue-bearing paper. Zero-issue papers are omitted, and papers
with more than 100 hidden issues are filtered before plotting.}
\label{fig:issue_count_histograms}
\end{figure*}

\Cref{fig:issue_count_histograms} breaks down the issue-count distribution from~\Cref{fig:dataset-dissection} by conference and track.
During data preprocessing, we deliberately excluded repositories with more than 100 issues, as collecting issues from such repositories would require excessive GitHub API requests.
It shows that the long-tailed pattern is not driven by a single split: in every venue, most selected papers have a few human-reported reproducibility issues, while a small number of repositories attract many reports.

The skew in \Cref{fig:issue_count_histograms} motivates reporting both
issue-level and paper-level metrics. Issue-level match rates measure how much
human issue evidence is recovered, while paper-level match rates prevent a small
number of heavily discussed repositories from dominating the interpretation.

\subsection{Failure Taxonomy}
\label{app:failure-taxonomy}

Typical accepted problems include environment failures, missing artifacts,
broken scripts or configs, paper-repository mismatches, training or inference
blockers, evaluation failures, and result discrepancies. We categorize the
human issue evidence along two axes: (1) whether the failure is visible as a
crash or silent because the code runs but yields the wrong setup or result, and
(2) whether it affects startup/setup or a later training/evaluation stage. This
yields four categories:
\begin{itemize}
    \item \emph{Crash immediate}: the code fails at startup, e.g., broken imports, missing modules, or environment errors that prevent any run.
    \item \emph{Crash late}: the code crashes mid-execution, e.g., shape mismatches, OOM, or runtime errors during training or evaluation.
    \item \emph{Silent wrong setup}: the released setup does not match the paper's intended configuration, artifact scope, checkpoint, data provenance, or evaluation path, even when individual scripts execute without error.
    \item \emph{Silent wrong number}: the code runs to completion but produces results that disagree with the paper's reported numbers.
\end{itemize}

\section{Additional Evaluation Results}
\label{app:additional-results}

\subsection{All-Finding Conference Comparison}
\label{app:all-finding-conference-comparison}

\Cref{tab:conference-comparison-all} repeats the conference comparison in
\Cref{tab:conference-comparison-10} after removing the top-10 reporting budget.
It communicates that the main top-10 results capture nearly all of the matched
issues: allowing all findings changes the reported rates only minimally.

\begin{table*}[t]
\centering
\begin{tabular}{lccccccc}
\toprule
Conference
& \multicolumn{2}{c}{Issue Match}
& \multicolumn{2}{c}{Paper Match (Any)}
& \multicolumn{2}{c}{Paper Match (All)}
& Avg. Cost \\
\cmidrule(lr){2-3}\cmidrule(lr){4-5}\cmidrule(lr){6-7}
& EM & SM
& EM & SM
& EM & SM
& \\
\midrule
NeurIPS 2022 main
& 23.8\% & 57.4\%
& 61.2\% & 89.7\%
& 15.2\% & 41.4\%
& \$1.28 \\
NeurIPS 2022 DB
& 18.2\% & 51.0\%
& 60.2\% & 96.6\%
& 10.2\% & 31.8\%
& \$1.50 \\
NeurIPS 2024 main
& 24.0\% & 56.4\%
& 61.0\% & 93.3\%
& 22.7\% & 51.1\%
& \$1.30 \\
NeurIPS 2024 DB
& 12.2\% & 47.3\%
& 46.7\% & 87.6\%
& 10.5\% & 33.3\%
& \$1.53 \\
ICLR 2026
& 25.4\% & 58.1\%
& 59.7\% & 89.7\%
& 34.8\% & 65.6\%
& \$1.26 \\
\bottomrule
\end{tabular}
\caption{\textbf{All-Finding Conference Comparison.} Diagnostic all-finding conference comparison for Codex with GPT-5.5,
using the same metrics as in \Cref{tab:conference-comparison-10} but without the top-10 cutoff.}
\label{tab:conference-comparison-all}
\end{table*}

\subsection{Token Usage and Cost}
\label{app:token-usage-cost}

\Cref{tab:codex_gpt55_cost_breakdown} expands the token and cost columns in
\Cref{tab:conference-comparison-10}. The table shows that Datasets and
Benchmarks tracks require more tokens and cost more per run than the
corresponding main-conference splits, consistent with larger or more complex
artifact structures.

\begin{table*}[t]
\centering
\begin{tabular}{lrrrrr}
\toprule
Conference
& Uncached Input
& Cached Input
& Output
& Total Tokens
& Cost \\
\midrule
NeurIPS 2022 main & 103,922 & 851,657  & 11,215 & 966,794   & \$1.28 \\
NeurIPS 2022 DB   & 118,001 & 1,095,238 & 12,136 & 1,225,375 & \$1.50 \\
NeurIPS 2024 main & 112,126 & 801,607  & 11,335 & 925,068   & \$1.30 \\
NeurIPS 2024 DB   & 126,678 & 1,019,991 & 13,029 & 1,159,698 & \$1.53 \\
ICLR 2026         & 98,834  & 832,592  & 11,586 & 943,011   & \$1.26 \\
\bottomrule
\end{tabular}
\caption{\textbf{Token Usage and Cost.} Codex with GPT-5.5 token usage and estimated cost per evaluated paper-snapshot pair, averaged within each conference. Total Tokens sums uncached input, cached input, and output tokens.}
\label{tab:codex_gpt55_cost_breakdown}
\end{table*}

\subsection{Ablation on Output Ordering}
\label{app:output-ordering-ablation}

\begin{figure*}[t]
    \centering
    \includegraphics[width=0.495\linewidth]{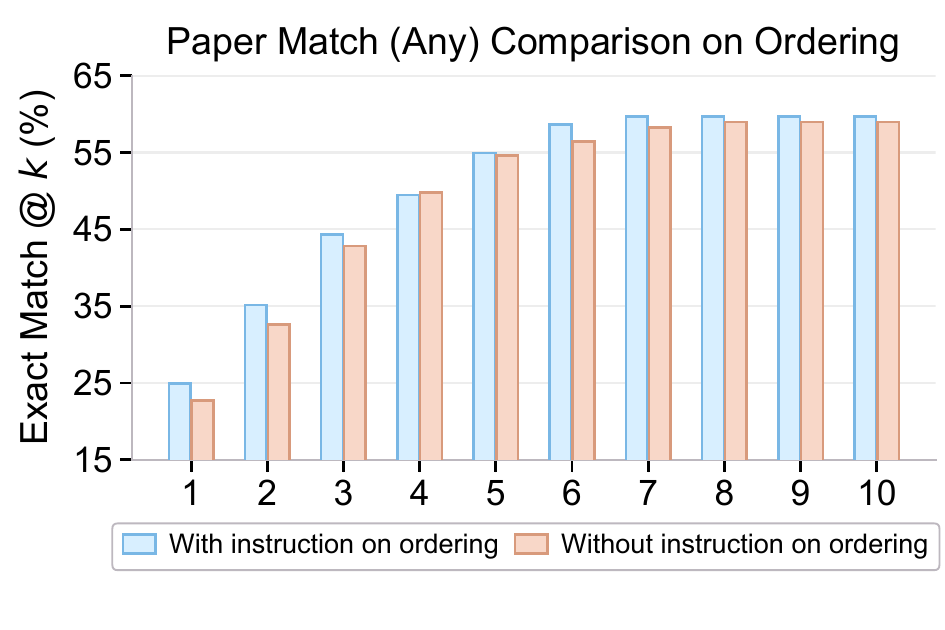}
    \includegraphics[width=0.495\linewidth]{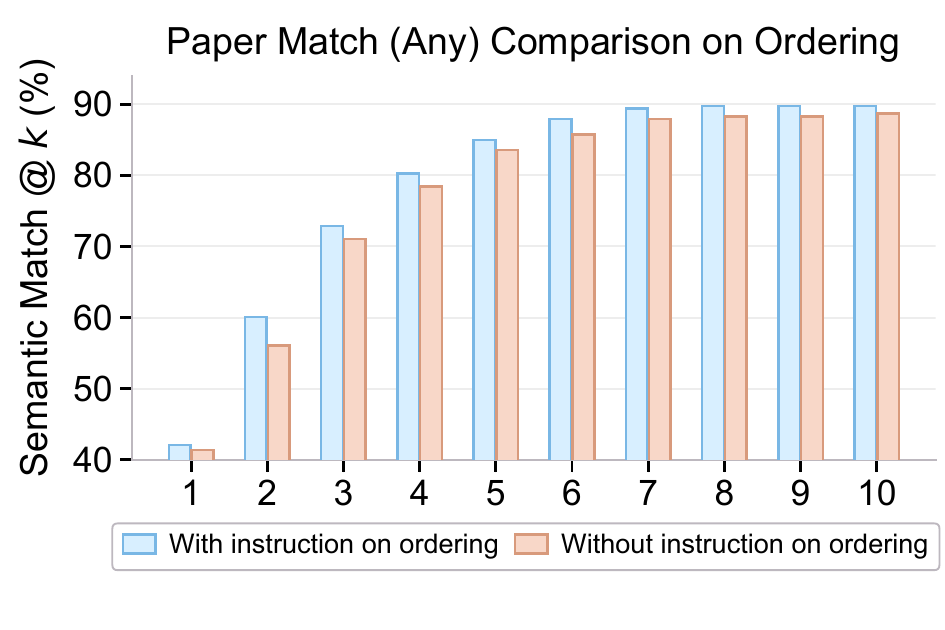}
    \vspace{-10mm}
    \caption{Ablation on the ordering instruction. We compare top-$k$ paper-level any-issue match rates with and without instructing the agent to order findings by the estimated likelihood of being encountered and reported by human users. Left: exact match. Right: semantic match.}
    \label{fig:topk-ablation-ordering}
\end{figure*}

We instruct the model to output potential issues in descending order of the
estimated likelihood that the issue would be encountered and reported by human
users. In \Cref{fig:topk-ablation-ordering}, we ablate this design choice by
comparing the top-$k$ paper-level any-issue match rate with and without the
ordering instruction. The ordering instruction \textbf{mainly improves the
ranking quality under small reporting budgets}. For both exact and semantic
matching, the model with the ordering instruction generally achieves higher
match rates when only the first few findings are considered, indicating that it
places human-reported blockers earlier in the report. As $k$ increases, the two
curves become close, especially when $k\geq 8$. This suggests that the
instruction does not substantially change the total set of issues the agent is
able to identify; rather, it helps prioritize the findings that are more likely
to correspond to real user-reported reproduction blockers. This makes the
ordered report more suitable for human triage under a limited inspection budget.

\subsection{Run-to-Run Variance}
\label{app:run-variance}

Agent outputs are stochastic, so a natural question is whether the match rates in \Cref{tab:model-comparison-10} are stable across repeated runs or reflect the luck of a single sample. To quantify this, we repeat the full \ourmethod pipeline three times, independently, for the two Claude Code and Codex configurations with the strongest model in each agent family, namely Claude Opus 4.7 and GPT-5.5. The first run of each configuration corresponds to the numbers reported in \Cref{tab:model-comparison-10}. \Cref{tab:run-variance} lists each of the three runs separately together with their mean and sample standard deviation.

\textbf{Run-to-run variance is small across every metric.} The standard deviation is at most $2.7$ percentage points and is typically below $1.5$ points. Crucially, the run-to-run spread is much smaller than the gaps between configurations and between EM and SM in \Cref{tab:model-comparison-10}. Therefore, none of the qualitative conclusions in \Cref{sec:exp}, including the large EM--SM gap, the high paper-level any-issue coverage near 90\% under semantic matching, and the ranking of configurations, depend on the particular sampled run, verifying the robustness of our claims.

\begin{table*}[t]
\centering
\resizebox{\linewidth}{!}{%
\begin{tabular}{@{}lccccccc@{}}
\toprule
Configuration & Run
& \multicolumn{2}{c}{Issue Match}
& \multicolumn{2}{c}{Paper Match (Any)}
& \multicolumn{2}{c}{Paper Match (All)} \\
\cmidrule(lr){3-4}\cmidrule(lr){5-6}\cmidrule(lr){7-8}
& & EM@10 & SM@10
& EM@10 & SM@10
& EM@10 & SM@10 \\
\midrule
\multirow{4}{*}{\shortstack[l]{Claude Opus 4.7\\(Claude Code)}}
& Run 1
& 21.7\% & 54.1\% & 52.0\% & 86.5\% & 27.8\% & 57.1\% \\
& Run 2 & 23.3\% & 55.4\% & 54.6\% & 88.3\% & 30.0\% & 56.8\% \\
& Run 3 & 23.1\% & 55.5\% & 53.5\% & 89.0\% & 28.6\% & 57.5\% \\
\cmidrule(l){2-8}
& Mean\scriptsize{$\pm$std}
& 22.7\%\scriptsize{$\pm$0.85\%} & 55.0\%\scriptsize{$\pm$0.76\%}
& 53.4\%\scriptsize{$\pm$1.29\%} & 87.9\%\scriptsize{$\pm$1.32\%}
& 28.8\%\scriptsize{$\pm$1.12\%} & 57.1\%\scriptsize{$\pm$0.37\%} \\
\midrule
\multirow{4}{*}{\shortstack[l]{GPT-5.5\\(Codex)}}
& Run 1
& 25.4\% & 58.1\% & 59.7\% & 89.7\% & 34.8\% & 65.6\% \\
& Run 2 & 24.5\% & 56.4\% & 58.2\% & 89.0\% & 33.0\% & 61.5\% \\
& Run 3 & 25.2\% & 59.9\% & 57.1\% & 90.5\% & 33.3\% & 66.7\% \\
\cmidrule(l){2-8}
& Mean\scriptsize{$\pm$std}
& 25.0\%\scriptsize{$\pm$0.50\%} & 58.1\%\scriptsize{$\pm$1.78\%}
& 58.4\%\scriptsize{$\pm$1.29\%} & 89.7\%\scriptsize{$\pm$0.73\%}
& 33.7\%\scriptsize{$\pm$0.97\%} & 64.6\%\scriptsize{$\pm$2.70\%} \\
\bottomrule
\end{tabular}
}
\caption{\textbf{Run-to-run variance.} Three independent runs on the ICLR 2026 set, using the same metrics as \Cref{tab:model-comparison-10}. Run 1 of each
configuration is the run reported in \Cref{tab:model-comparison-10}. The small
standard deviations indicate that the reported match rates are stable across
runs.}
\label{tab:run-variance}
\end{table*}

\subsection{Human Validation of Alignment Judge Labels}
\label{app:human-judge-validation}

Because \ourmethod uses an LLM judge to compare agent findings against real GitHub issues, we additionally perform a human validation study to check the quality of the judge.
We sample 150 finding-issue pairs where the findings are generated by Codex with GPT-5.5, with 50 examples from each automatic label: exact match, semantic match, and none.
For each pair, a human annotator is shown the agent-predicted issue, the candidate GitHub issue, and the judge rationale with the automatic label hidden, and assigns one of the same three labels.

Among the 50 pairs labeled EM by the judge, the human annotator labels 44 as EM, indicating high agreement under the strictest label. Among the 100 broadly matched pairs (including both EM and SM), the human annotator labels 84 as matched and 16 as none. Conversely, among the 50 pairs labeled None by the judge, the human annotator labels 44 as None and 6 as SM.
For the matched-vs.-unmatched decision, the human annotation agrees with the LLM judge on 128 of 150 examples ($85.3\%$). 
In addition, among the 90 examples that the human annotator labels as matched, the judge
recovers 84.
Thus, the match decision has $84.0\%$ precision and $93.3\%$ recall within the validation sample.
Therefore, the human labels support the reliability of the LLM judge.

\subsection{Analysis on Repositories without Reported Issues}

As a complementary analysis, we audit three ICLR 2026 repositories for which our collection found no open or closed GitHub issues. Codex with GPT-5.5 produces 18 findings; manual inspection finds 14 supported concerns, 3 genuine but overstated concerns, and 1 unsupported finding. We point out that repositories without reported issues are not necessarily clean negatives due to sparse usage/reporting. This analysis also provide preliminary evidence that the agent can surface plausible and manually verifiable reproducibility concerns in repositories outside the issue-positive benchmark subset. 

\section{Agent Prediction Taxonomy}
\label{app:agent-prediction-taxonomy}

The following figures apply the same taxonomy breakdown applied to human issues to agent predictions. Together, they show how the breakdown of agent findings compares with that of the human-raised GitHub issues, and how this distribution changes when predictions are restricted to exact or semantic matches.

\begin{figure*}[!t]
\centering
    \includegraphics[width=0.85\linewidth]{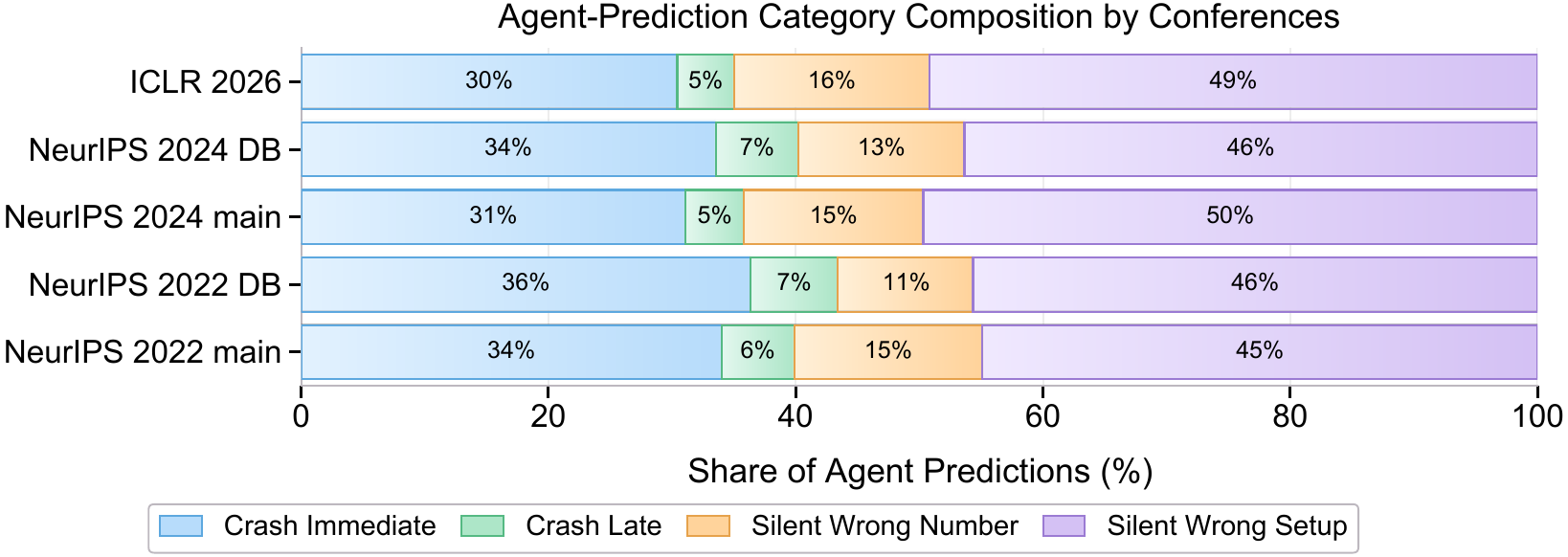}
\caption{\textbf{Agent Prediction Taxonomy Analysis.} The distribution of agent-predicted issue categories across conference splits.}
\label{fig:agent-prediction-category-all}
\end{figure*}

\Cref{fig:agent-prediction-category-all} shows the category distribution across
all agent predictions: from static inspection, the agent most commonly surfaces issues related to \emph{silent wrong setup}, followed by \emph{crash immediate}, whereas both \emph \emph{silent wrong number} and \emph{crash late} are less commonly predicted. 
This highlights that along the two axes, the crash vs. silent distinction is much less salient in agent predictions than the setup vs. later-stage distinction. While it consistently identifies slightly more silent failures than crash failures, it predominantly predicts setup/immediate failures compared to number/late failures.

\begin{figure*}[!t]
\centering
    \includegraphics[width=0.85\linewidth]{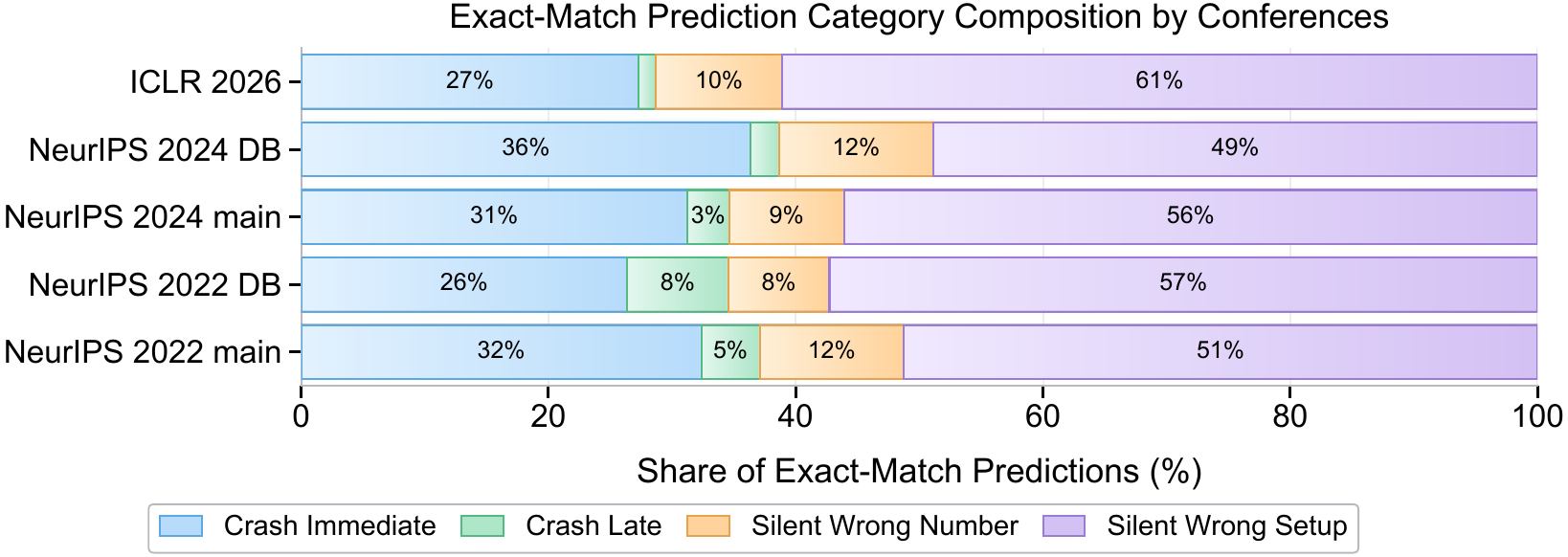}
\caption{\textbf{Exact-Match Agent Prediction Taxonomy Analysis.} The distribution of exact-match agent-predicted issue categories across conference splits.}
\label{fig:agent-prediction-category-exact-match}
\end{figure*}

\begin{figure*}[!t]
\centering
    \includegraphics[width=0.85\linewidth]{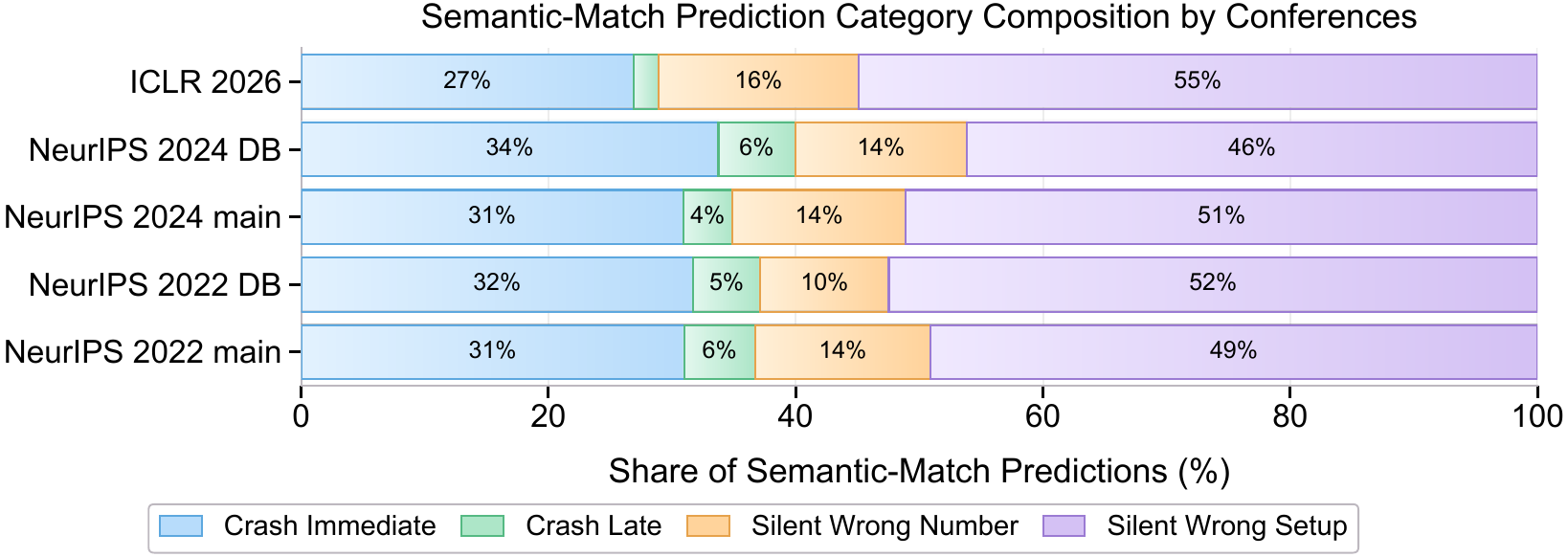}
\caption{\textbf{Semantic-Match Agent Prediction Taxonomy Analysis.} The distribution of semantic-match agent-predicted issue categories across conference splits.}
\label{fig:agent-prediction-category-semantic-match}
\end{figure*}

\Cref{fig:agent-prediction-category-exact-match} restricts the distribution to
predictions that exactly match a hidden human issue, and \Cref{fig:agent-prediction-category-semantic-match} restricts to
semantically matched predictions. In general, these show a similar pattern as \Cref{fig:agent-prediction-category-all}.

\section{Prompt Templates}
\label{app:prompt-templates}

\subsection{Agent Prompt}
\label{app:discovery-prompt}

\begin{Verbatim}[fontsize=\scriptsize,breaklines=true,breakanywhere=true]
Generate a blind, static, no-execution reproducibility issue discovery result
for the provided research paper PDF and repository snapshot.

Blind audit rules:
- Use only the provided paper PDF and the local repository files in this
  workspace.
- Do not inspect GitHub issues, pull requests, discussions, comments, commit
  history, remote branches/tags, internet search, external websites, hidden
  benchmark metadata, or any repository state outside this snapshot.
- Do not run setup, tests, scripts, notebooks, downloads, training, inference,
  evaluation, or benchmark commands.

Goal:
Identify GitHub-issue-style reproduction blockers that a real user would be
likely to report when trying to reproduce the paper from this repository
snapshot.

Focus on concrete static evidence:
- missing or ambiguous data/checkpoint/model artifacts
- broken or incomplete installation/dependency instructions
- missing scripts/configs/paths referenced by the paper or README
- paper/repository mismatches
- evaluation/training workflow gaps
- result/metric provenance gaps
- code/config patterns that would cause reproduction-time crashes or silent
  wrong numbers

Prefer atomic findings: each finding should describe one user-facing blocker,
its trigger, root cause, static evidence, and impact on reproducing the paper.

Order findings by expected human-report likelihood, highest first. Prioritize
issues a normal reproducer would actually hit and open as a GitHub issue:
immediate blockers, documented workflow failures, missing required artifacts,
misleading instructions, common setup/data/model paths, and result mismatches.
Put broad audit observations, speculative risks, and low-actionability issues
later or omit them.

Use these finding categories when possible:
- environment_setup
- dependency_version
- data_access
- model_weights
- training_reproduction
- evaluation_reproduction
- runtime_error
- hardware_backend
- api_service
- result_mismatch
- documentation_gap
- paper_repo_mismatch

Write exactly one required artifact:
- ./findings.json

Required ./findings.json schema:

{{
  "reproducibility_assessment": "...",
  "findings": [
    {{
      "finding_id": "F01",
      "rank_reason": "Why this issue is likely to be reported by a human reproducer before lower-ranked findings.",
      "severity": "critical|high|medium|low",
      "stage": "install/environment|data/model acquisition|training/fine-tuning|inference/demo|evaluation/metrics|result comparison/provenance",
      "category": "environment_setup|dependency_version|data_access|model_weights|training_reproduction|evaluation_reproduction|runtime_error|hardware_backend|api_service|result_mismatch|documentation_gap|paper_repo_mismatch",
      "issue_title": "...",
      "user_symptom": "...",
      "trigger_context": "...",
      "root_cause": "...",
      "evidence": [
        {{
          "path": "...",
          "line_refs": "...",
          "evidence_type": "file|command|missing_artifact|documentation|config|paper_repo_mismatch",
          "summary": "..."
        }}
      ],
      "impact": "...",
      "affected_claims_or_metrics": ["..."],
      "reproducibility_blocker": true,
      "confidence": "high|medium|low"
    }}
  ],
  "missing_or_ambiguous_artifacts": ["..."]
}}

The JSON finding fields should contain enough information for a separate judge
to compare them against hidden GitHub issues by symptom, trigger, root cause,
evidence, and impact. If you find no concrete reproduction blockers, write an
empty findings array.
\end{Verbatim}

\subsection{Judge Prompt}
\label{app:alignment-judge-prompt}

\begin{Verbatim}[fontsize=\scriptsize,breaklines=true,breakanywhere=true]
You are an alignment judge for a reproducibility benchmark.

You will receive exactly one file: ./task.json. It contains hidden GitHub issue
benchmark cases and blind agent findings for the same paper/repository snapshot.

Allowed information:
- The hidden issue cases in ./task.json.
- The blind agent findings in ./task.json.

Forbidden information:
- Do not browse the web.
- Do not inspect GitHub issues, pull requests, commits, discussions, release
  pages, external websites, or repository files.
- Do not run commands other than reading ./task.json and writing outputs.
- Do not infer from issue numbers, URLs, PR numbers, or commit hashes.

Leakage policy:
- The discovery agent did not see hidden issue text, comments, PRs, commits, or
  gold answers. Judge only after discovery is complete.
- For closed_gold cases, any provenance-only fix evidence is optional
  adjudication support. Do not require the blind finding to mention patch files
  or post-fix implementation details unless the hidden issue itself identifies
  those details as the user-facing reproduction problem.

Your job:
For each hidden issue, decide whether the best matching agent finding is:
- "exact": substantially the same user-facing reproduction problem.
  It should align on the main symptom and at least most of trigger/workflow,
  root cause, involved artifact, and impact. Wording can differ.
- "semantic": a related blocker or the same workflow area is identified, but
  with incomplete specificity. The Codex finding and GitHub issue share a
  concrete reproduction context, but do not describe the exact same blocker.
- "none": no agent finding would be useful evidence that the agent discovered
  this hidden issue.

Comparison dimensions:
- user-facing symptom
- trigger path / workflow
- root cause
- involved artifact, file, command, config, dataset, model, or metric
- impact on reproducing the paper

Also judge each agent finding against all hidden issues so downstream summaries
can estimate over-reporting. A finding may match zero, one, or multiple hidden
issues if it genuinely covers multiple human reports.

Write exactly one required artifact:
- ./result.json

Required JSON schema:
{{
  "task_id": "{task['task_id']}",
  "snapshot_group_id": "{task['snapshot_group_id']}",
  "issue_alignments": [
    {{
      "hidden_issue_id": "string from task.hidden_issues[].hidden_issue_id",
      "target": "open|closed_gold|closed_silver",
      "issue_number": 0,
      "best_label": "exact|semantic|none",
      "best_finding_id": "string or null",
      "best_finding_index": 0,
      "dimension_labels": {{
        "user_facing_symptom": "exact|semantic|none",
        "trigger_path_or_workflow": "exact|semantic|none",
        "root_cause": "exact|semantic|none",
        "involved_artifact": "exact|semantic|none",
        "impact": "exact|semantic|none"
      }},
      "rationale": "short explanation grounded in both the hidden issue and finding",
      "matched_evidence": ["short strings"]
    }}
  ],
  "finding_alignments": [
    {{
      "finding_id": "string",
      "finding_index": 0,
      "best_label": "exact|semantic|none",
      "matched_hidden_issue_ids": ["zero or more hidden issue ids"],
      "matched_targets": ["open|closed_gold|closed_silver"],
      "rationale": "short explanation"
    }}
  ],
  "notes": ["optional short notes about ambiguity"]
}}

Use only the three allowed labels exactly as written. If there are no hidden
issues or no findings, still write valid arrays.
\end{Verbatim}


\section{Agent Output Examples}
\label{app:alignment-examples}

In this section, we provide concrete examples on exact match,
semantic match, and no match.

\subsection{Exact Match Example}
\begin{humanissuebox}
\small
{\bfseries real data test}\hfill\ghpill{human report}

\tcblower

\par\smallskip\noindent{\scriptsize\bfseries\href{https://github.com/caiyuanhao1998/MST/issues/5}{GitHub issue}}\par
\noindent
Is the test command wrong?
\par

\par\smallskip\noindent{\scriptsize\bfseries\href{https://github.com/caiyuanhao1998/MST/issues/5\#issuecomment-1808763278}{GitHub discussion}}\par
\noindent
\texttt{@dongruoyu1} I believe so (``test in 5.2 is incorrect''). At least the script didn't work.
5.2 Real Test script looks like a direct copy of the simulation one but doesn't work on real data.

e.g.
\begin{quote}
{\color{blue!55!black}\small\ttfamily
python test.py --template mst\_s --outf ./exp/mst\_s/\\
--method mst\_s --pretrained\_model\_path ./model\_zoo/mst/mst\_s.pth
\par}
\end{quote}

But the \textbf{format} of real measurements is different from simulation so the real test code could not load it.
When I ran the aforementioned code, I got this error:
\href{https://github.com/caiyuanhao1998/MST/issues/36}{\#36}
\par
\end{humanissuebox}

\begin{agentfindingbox}
\small
{\bfseries Documented real-test commands do not match
\texttt{real/test\_code/test.py} and use an unreachable hard-coded
checkpoint.}\hfill\ghpill[blue]{agent finding}

\tcblower

\issuefield{User symptom}{Running the README's real testing command with
\texttt{--template}, \texttt{--outf}, \texttt{--method}, and
\texttt{--pretrained\_model\_path} is rejected by \texttt{argparse}, and editing
around that still leads to missing \texttt{./Data} inputs and a hard-coded
\texttt{/data/lj/...} checkpoint path.}

\issuefield{Trigger context}{Follow README section ``4. Real Experiement'' and run
a listed command from \texttt{repo/real/test\_code/}, for example:
\begin{quote}
{\color{blue!55!black}\small\ttfamily
python test.py --template mst\_s --outf ./exp/mst\_s/\\
--method mst\_s --pretrained\_model\_path ./model\_zoo/mst/mst\_s.pth
\par}
\end{quote}}

\issuefield{Root cause}{\texttt{repo/real/test\_code/test.py} is a standalone
stale script. It defines a different argument interface, hard-codes CUDA device
0, hard-codes default data under \texttt{./Data}, ignores the README
\texttt{model\_zoo} path, and loads a developer-local absolute checkpoint path.}

\issuefield{Impact}{The documented real-data inference path cannot be used to
reproduce the paper's real reconstruction results without rewriting the script
and supplying undocumented local artifacts.}
\end{agentfindingbox}

This is an exact match: the LLM predicts\textbf{ exactly the same trigger command} as the one discussed in the GitHub issue, highlighted in {\color{blue!55!black}blue} above, and recovers the same
user-facing workflow and practical blocker.

\subsection{Semantic Match Example}

\begin{humanissuebox}
\small
{\bfseries Inference of DAT model}\hfill\ghpill{human report}

\tcblower

\par\smallskip\noindent{\scriptsize\bfseries\href{https://github.com/alibaba/easyrobust/issues/12}{GitHub issue}}\par
\noindent
Hi, thank you for your work of `Discrete Adversarial Training'

I was trying to use the pre-tained model for inference. The [ViT-B/16] works
pretty well. However, the [MAE-ViT-H] doesn't work. Could you tell the
difference between this two model when testing them on the custom images?

For the [MAE-ViT-H], I used \texttt{vit\_huge\_patch14\_224} instead of
\texttt{mae\_vit\_huge\_patch14} to initialize the model in \texttt{timm}.
(the later can not be found in \texttt{timm} model zoo.)

I normalize the input image with
\texttt{T.Normalize((0.485, 0.456, 0.406), (0.229, 0.224, 0.225))}

Thank you !
\par
\end{humanissuebox}

\begin{agentfindingbox}
\small
{\bfseries No training recipe is provided for the headline MAE-H DAT
fine-tuning result.}\hfill\ghpill[blue]{agent finding}

\tcblower

\issuefield{User symptom}{A user trying to reproduce the reported 31.40
ImageNet-C mCE / 32.77 Stylized-ImageNet MAE-H result can only download an
already trained checkpoint; they cannot reproduce the fine-tuning run from
documented commands or local configs.}

\issuefield{Trigger context}{Use the paper abstract/Table 1 or DAT README
pretrained-weights table and look for the corresponding MAE-H fine-tuning
command/config.}

\issuefield{Root cause}{The only DAT training command in the README trains
\texttt{vit\_base\_patch16\_224} and does not specify
\texttt{mae\_vit\_huge\_patch14}, a MAE initialization checkpoint,
MAE-specific input size/crop settings, or fine-tuning hyperparameters. The
MAE-H section is evaluation-only.}

\issuefield{Impact}{The paper's most prominent reproducibility target cannot be
regenerated from the repository snapshot; users can only verify a released
final checkpoint if the external URL is available.}
\end{agentfindingbox}

This is a semantic match: both sides identify the same MAE-H DAT model artifact
as a reproduction-critical object, but the human issue is an inference-time
model-name/import failure whereas the agent finding is about the missing
fine-tuning recipe.

\subsection{No Match Example}

\begin{agentfindingbox}
\small
{\bfseries DANN, ADDA, and SGADA training scripts import \texttt{wandb} but
requirements omit it.}\hfill\ghpill[blue]{agent finding}

\tcblower

\issuefield{User symptom}{After installing the documented requirements for
UFLD-DANN, UFLD-ADDA, or UFLD-SGADA, running \texttt{train.py} would fail at
startup with \texttt{ModuleNotFoundError: No module named 'wandb'} unless
\texttt{wandb} was installed separately.}

\issuefield{Trigger context}{Follow any of the DANN/ADDA/SGADA READMEs:
\texttt{pip install -r requirements.txt}, then
\texttt{python train.py configs/...}.}

\issuefield{Root cause}{The train scripts import and use Weights \& Biases, but
the corresponding \texttt{requirements.txt} files list only
\texttt{opencv-python}, \texttt{tqdm}, \texttt{tensorboard}, \texttt{addict},
\texttt{sklearn}, and \texttt{pathspec}.}

\issuefield{Impact}{The installation instructions are incomplete for three
reported UDA baselines, blocking reproduction before any data or model logic
runs.}
\end{agentfindingbox}

The agent finding is still a plausible reproducibility issue. However, this kind of missing-dependency problem is often easy for users to resolve locally without opening a GitHub issue. No existing GitHub issues match this agent finding.

\end{document}